\documentclass{article} 
\usepackage{iclr2023_conference,times}

\usepackage{amsmath,amsfonts,bm}

\def\eqref#1{equation~\ref{#1}}

\def\1{\bm{1}}

\DeclareMathAlphabet{\mathsfit}{\encodingdefault}{\sfdefault}{m}{sl}
\SetMathAlphabet{\mathsfit}{bold}{\encodingdefault}{\sfdefault}{bx}{n}

\definecolor{linkColor}{rgb}{0.18,0.39,0.62}
\usepackage[pagebackref=true,breaklinks=true,colorlinks,citecolor=linkColor]{hyperref}

\usepackage{url}
\usepackage{wrapfig}
\usepackage{graphicx}
\usepackage{multirow}
\usepackage{makecell}
\usepackage{booktabs}
\usepackage{amssymb}
\usepackage{longtable}
\usepackage{xcolor}
\usepackage{colortbl}
\usepackage{hhline}
\usepackage{color}
\usepackage[misc]{ifsym}
\newcommand{\tablestyle}[2]{\setlength{\tabcolsep}{#1}\renewcommand{\arraystretch}{#2}\centering\footnotesize}
\title{Vision Transformer Adapter for \\ Dense Predictions}

\iclrfinalcopy
\author{
    Zhe Chen$^{1,2}$\thanks{Equal contribution. ~\textsuperscript{\Letter}Corresponding authors.}~, Yuchen Duan$^{2,3*}$, Wenhai Wang$^2$\textsuperscript{\Letter}, Junjun He$^2$, \\
    ~\textbf{Tong Lu$^1$\textsuperscript{\Letter}, Jifeng Dai$^{2,3}$, Yu Qiao$^{2}$}\\
    ~$^1$Nanjing University, $^2$Shanghai AI Laboratory, $^3$Tsinghua University \\
    ~\small{\texttt{czcz94cz@gmail.com}, \texttt{\{duanyuchen,wangwenhai,hejunjun\}@pjlab.org.cn}}\\
    ~\small{\texttt{lutong@nju.edu.cn}, \texttt{\{daijifeng,qiaoyu\}@pjlab.org.cn}}
}

\begin{document}

\def\ie{\textit{i.e.}}
\def\eg{\textit{e.g.}}
\def\etc{etc}
\def\etal{\textit{et al.}}

\maketitle

\vspace{-1em}
\begin{abstract}
This work investigates a simple yet powerful dense prediction task adapter for Vision Transformer (ViT).
Unlike recently advanced variants that incorporate vision-specific inductive biases into their architectures, the plain ViT suffers inferior performance on dense predictions due to weak prior assumptions.
To address this issue, we propose the ViT-Adapter, which allows plain ViT to achieve comparable performance to vision-specific transformers.
Specifically, the backbone in our framework is a plain ViT that can learn powerful representations from large-scale multi-modal data.
When transferring to downstream tasks, a \textbf{pre-training-free adapter} is used to introduce the image-related inductive biases into the model, making it suitable for these tasks.
We verify ViT-Adapter on multiple dense prediction tasks, including object detection, instance segmentation, and semantic segmentation.
Notably, without using extra detection data, our ViT-Adapter-L yields state-of-the-art \textbf{60.9} box AP and \textbf{53.0} mask AP on COCO test-dev.
We hope that the ViT-Adapter could serve as an alternative for vision-specific transformers and facilitate future research.
Code and models will be released at \small{\url{https://github.com/czczup/ViT-Adapter}}.
\end{abstract}

\vspace{-2mm}
\begin{figure}[h]
    \centering
    \includegraphics[width=0.85\linewidth]{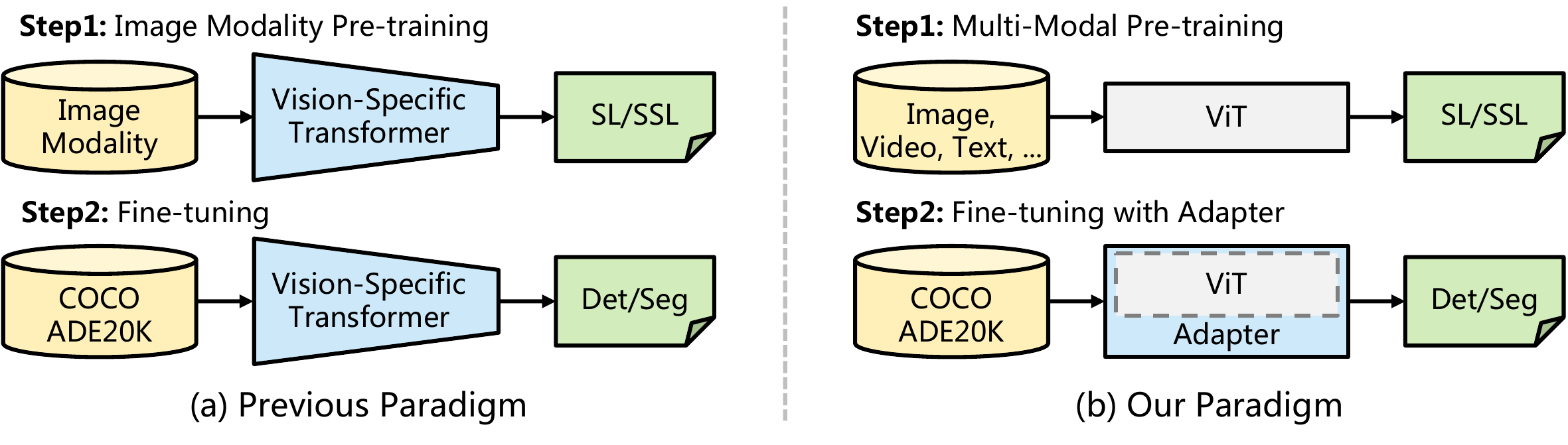}
    \vspace{-0.5em}
    \caption{
    \textbf{Previous paradigm \emph{vs.} our paradigm.} 
    (a) Previous paradigm designs vision-specific models and pre-trains on large-scale image datasets via supervised or self-supervised learning and then fine-tunes them on downstream tasks.
    (b) We propose a pre-training-free adapter to close the performance gap between plain ViT~\citep{dosovitskiy2020image} and vision-specific transformers (\eg, Swin~\citep{liu2021swin}) for dense prediction tasks.
    Compared to the previous paradigm, our method preserves the flexibility of ViT and thus could benefit from advanced multi-modal pre-training.
    }
    \label{fig:paradigm}
\end{figure}
\vspace{-2.5mm}

\section{Introduction}
\label{sec:introduction}
Recently, transformers have witnessed remarkable success in a broad range of computer vision fields.
Benefiting from the dynamic modeling capability and the long-range dependence of the attention mechanism, various vision transformers~\citep{dosovitskiy2020image,chen2021simple,han2021transformer,li2021localvit,wu2022p2t} soon rose in many computer vision tasks such as object detection and semantic segmentation, surpassing CNN models and reaching state-of-the-art performance.
These models are mainly divided into two families, \ie~the plain ViT \citep{dosovitskiy2020image, touvron2021training}, and its hierarchical variants \citep{dong2021cswin,liu2021swin,wang2021pyramid, wang2021pvtv2}.
In general, the latter can produce better results and is believed to introduce vision-specific inductive biases into their architectures by using local spatial operations.

\begin{wrapfigure}{r}{0.5\textwidth}
    \includegraphics[width=0.99\linewidth]{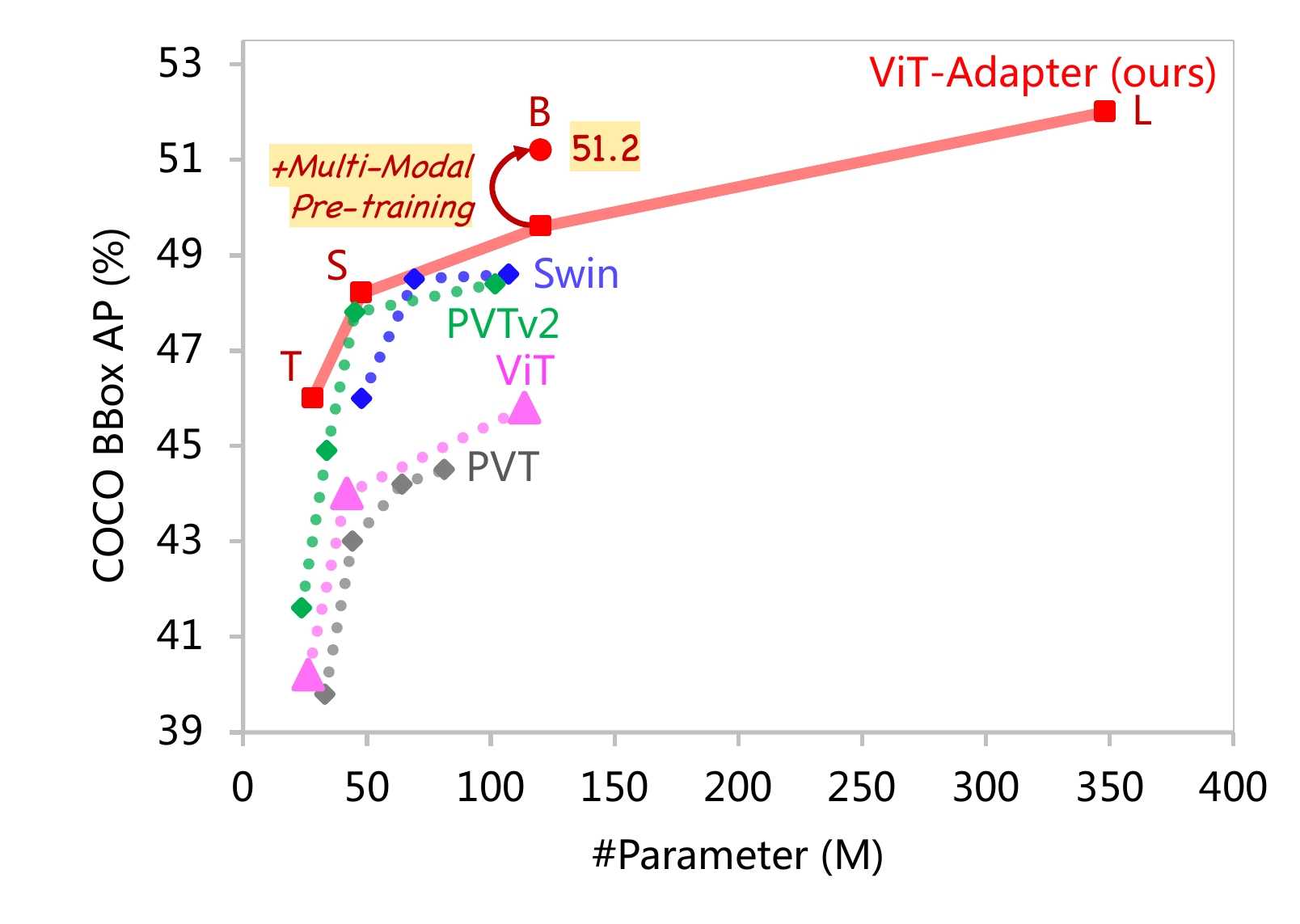}
    
    \vspace{-39.0mm}\hspace{30.2mm}
    \resizebox{0.26\columnwidth}{!}{\tablestyle{2pt}{1}
    	\renewcommand\arraystretch{0.55}
    	\begin{tabular}[b]{l|c|c}
	\renewcommand{\arraystretch}{0.1}
	Method & \#Param & AP$\rm ^b$ \\
	\hline
	PVTv2-B1   & 33.7M  & 44.9 \\
	ViT-T      & 26.1M  & 40.2 \\
	\rowcolor{gray!20}
	ViT-Adapter-T (ours)            & 28.1M  & 46.0 \\
	\hline
	PVTv2-B2   & 45.0M  & 47.8 \\
	Swin-T     & 47.8M  & 46.0 \\
	ViT-S & 43.8M  & 44.0 \\
	\rowcolor{gray!20}
	ViT-Adapter-S (ours)            & 47.8M  & 48.2 \\
	\hline
	Swin-B     & 107.1M & 48.6 \\
	ViT-B      & 113.6M & 45.8 \\
	\rowcolor{gray!20}
	ViT-Adapter-B (ours)            & 120.2M & 49.6 \\
	\rowcolor{yellow!15}
	ViT-Adapter-B$^\bigstar$(ours)   & 120.2M & 51.2 \\
	\hline
	ViT-L$^\dagger$      & 337.3M & 48.8 \\
	\rowcolor{gray!20}
	ViT-Adapter-L$^\dagger$ (ours)            & 347.9M & 52.1 \\
\end{tabular}
	}
    \vspace{9mm}
    \caption{
    \textbf{Object detection performance on COCO val2017 using Mask R-CNN.} 
    We see that the proposed ViT-Adapter brings significant improvements to plain ViTs.
    $^\bigstar$~indecates using multi-modal pre-trained ViT from~\citep{zhu2021uni}.
    Backbones pre-trained on ImageNet-22K are marked with $^\dagger$, otherwise ImageNet-1K.
    }
    \vspace{-3mm}
    \label{fig:param_ap}
\end{wrapfigure}

Nonetheless, the plain ViT (\ie, vanilla transformer) still has some nonnegligible advantages. 
A typical example lies in multi-modal pre-training~\citep{zhu2021uni,zhu2022uni,wang2022beit3}.
Stemming from the natural language processing (NLP) field, transformer has no assumption of input data. 
Equipping with different tokenizers, \eg~patch embedding~\citep{dosovitskiy2020image}, 3D patch embedding~\citep{liu2021video}, and token embedding~\citep{vaswani2017attention}, vanilla transformers such as \emph{plain ViT can use massive multi-modal data for pre-training}, including image, video, and text, which encourages the model to learn semantic-rich representations. 
However, the plain ViT has conclusive defects in dense predictions compared to vision-specific transformers. 
Lacking image-related prior knowledge results in slower convergence and lower performance, and thus plain ViTs are hard to compete with vision-specific transformers~\citep{huang2021shuffle,xie2021segformer,wang2021pvtv2} on dense prediction tasks.
Inspired by the adapters~\citep{houlsby2019parameter,stickland2019bert} in the NLP field, \emph{this work aims to develop an adapter to close the performance gap between the plain ViT and vision-specific backbones for dense prediction tasks.}

To this end, we propose the Vision Transformer Adapter (ViT-Adapter),
which is a \emph{pre-training-free} additional network that can efficiently adapt the plain ViT to downstream dense prediction tasks without modifying its original architecture.
Specifically, to introduce the vision-specific inductive biases into the plain ViT, we design three tailored modules for ViT-Adapter, including (1) a spatial prior module for capturing the local semantics (spatial prior) from input images, (2) a spatial feature injector for incorporating spatial prior into the ViT, and (3) a multi-scale feature extractor to reconstruct the multi-scale features required by dense prediction tasks.

As shown in Figure~\ref{fig:paradigm}, compared to the previous paradigm that pre-trains on large-scale image datasets (\eg, ImageNet~\citep{deng2009imagenet}) then fine-tunes on other tasks, our paradigm is more flexible. 
In our framework, the backbone network is a general-propose model (\eg, plain ViT) that can be pre-trained with not only images but also multi-modal data.
For the transfer learning of dense prediction tasks, we use a randomly initialized adapter to introduce the image-related prior knowledge (inductive biases) into the pre-trained backbone, making the model suitable for these tasks.
In this way, using ViT as the backbone, our framework achieves comparable or even better performance than vision-specific transformers such as Swin \citep{liu2021swin}.

Our main contributions are as follows:

$\bullet$ We explore a new paradigm to introduce vision-specific inductive biases into the plain ViT. It helps ViT achieve comparable performance to recent transformer variants~\citep{liu2021swin, wang2021pvtv2} with regular ImageNet pre-training and further benefits from multi-modal pre-training.

$\bullet$ We design a spatial prior module and two feature interaction operations, to inject the image prior without redesigning the architecture of ViT. They can supplement the missing local information and reorganize fine-grained multi-scale features for dense prediction tasks.

$\bullet$ We evaluate the ViT-Adapter on multiple challenging benchmarks, including COCO~\citep{lin2014microsoft} and ADE20K~\citep{zhou2017scene}.
As shown in Figure~\ref{fig:param_ap}, our models consistently achieve improved performance compared to the prior arts \emph{under the fair pre-training strategy}.
For instance, when using only ImageNet-1K pre-training, ViT-Adapter-B reports 49.6 box AP on COCO val, outperforming Swin-B by 1.0 points.
Benefiting from multi-modal pre-training~\citep{peng2022beitv2}, our ViT-Adapter-L yields 60.9 box AP, which is the best record on COCO test-dev without training on extra detection data such as Objects365~\citep{shao2019objects365}.

\vspace{-1em}
\section{Related Work}

\noindent \textbf{Transformers.}
In recent years, transformers have dominated various tasks across multiple modalities, such as natural language processing, computer vision, and speech recognition. 
The vanilla transformer \citep{vaswani2017attention} was initially proposed for machine translation and remains the state-of-the-art architecture for NLP tasks today.
ViT \citep{dosovitskiy2020image} is the first work to generalize the vanilla transformer to the image classification task without much modification. 
PVT \citep{wang2021pyramid} and Swin \citep{liu2021swin} introduce more vision-specific inductive biases by incorporating the pyramid structure from CNNs.
Afterward, Conformer~\citep{peng2021conformer} proposed the first dual network to combine CNN with transformer.
Recently, BEiT~\citep{bao2021beit} and MAE \citep{he2021masked} extended the scope of ViT to self-supervised learning with masked image modeling (MIM), demonstrating the powerful potential of the plain ViT architecture.
Many works~\citep{li2021benchmarking,zhu2021uni,zhu2022uni,wang2022beit3} have shown that designing vision-specific models is an important direction, but the general-propose architectures (\eg, plain ViT) are more flexible and essential for masked data modeling and multi-modal pre-training.
Therefore, we develop a pre-training-free adapter to introduce the image prior without modifying the architecture of ViT, preserving its flexibility and enjoying advanced multi-modal pre-training.
\begin{figure}[tbp]
    \centering
    \includegraphics[width=0.75\linewidth]{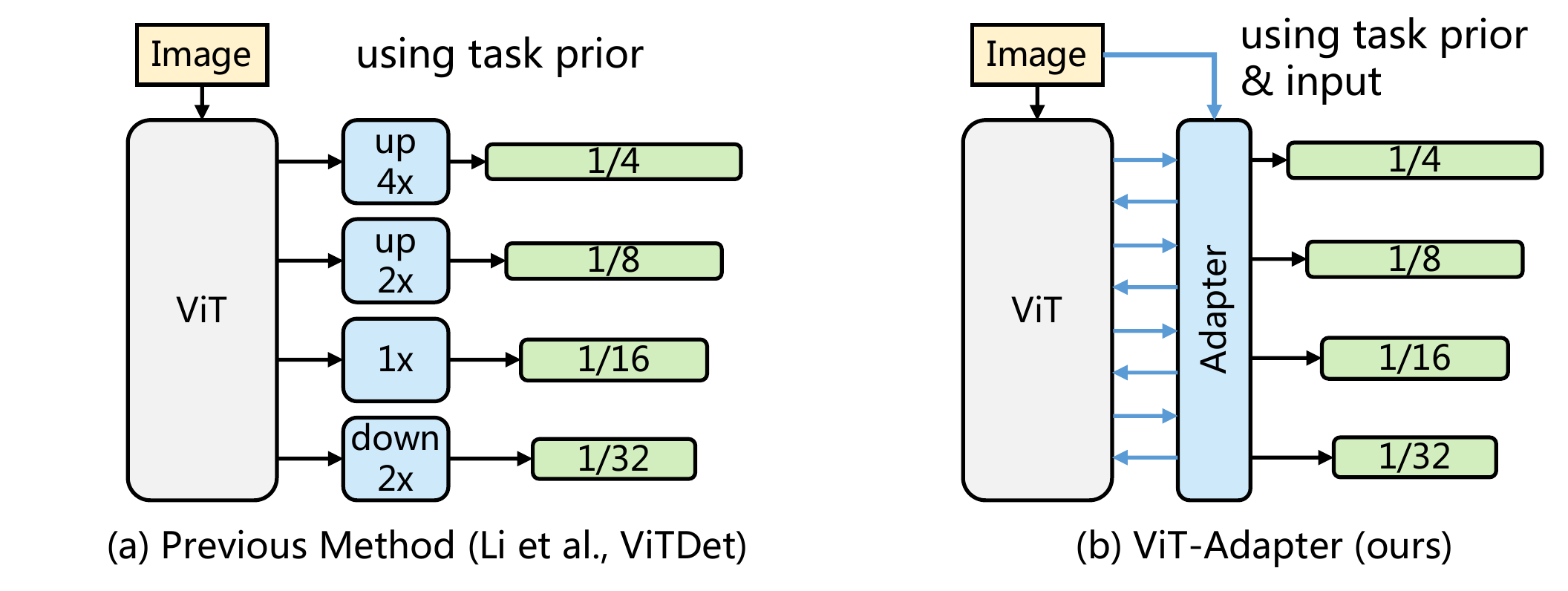}
    \vspace{-0.5em}
    \caption{
        \textbf{Overview of ViT-Adapter and two related approaches.}
        \citet{li2021benchmarking} and ViTDet~\citep{li2022exploring} build simple feature pyramid to adapt plain ViT for object detection, which only consider task prior.
        Differently, our adapter utilizes both task prior and the input image.
        }
    \label{fig:comparison}
\end{figure}

\noindent \textbf{Decoders for ViT.}
The architecture for dense prediction commonly follows an encoder-decoder pattern, in which the encoder generates rich features and the decoder aggregates and translates them to the final predictions.
Recently, illuminated by the global receptive fields of ViT, many works employ it as the encoder and design task-specific decoders.
SETR~\citep{zheng2021rethinking} is the first work to adopt ViT as the backbone and develop several CNN decoders for semantic segmentation.
Segmenter~\citep{strudel2021segmenter} also extends ViT to semantic segmentation, but differs in that it equips a transformer-based decoder.
DPT~\citep{ranftl2021vision} further applies ViT to the monocular depth estimation task via a CNN decoder and yields remarkable improvements.
In summary, these works improve the dense prediction performance of ViT by designing modality- and task-specific decoders, but remain ViT's weakness of single-scale and low-resolution representation.

\noindent \textbf{Adapters.}
To date, adapters have been widely used in the NLP field. 
PALs \citep{stickland2019bert} and Adapters \citep{houlsby2019parameter} introduce new modules in transformer encoders for task-specific fine-tuning, making the pre-trained model quickly adapt to downstream NLP tasks.
In the field of computer vision, some adapters have been proposed for incremental learning~\citep{rosenfeld2018incremental} and domain adaptation~\citep{rebuffi2017learning,rebuffi2018efficient}.
With the advent of CLIP~\citep{radford2021learning}, many CLIP-based adapters~\citep{gao2021clip,sung2021vl,zhang2021tip} were presented to transfer pre-trained knowledge to zero-shot or few-shot downstream tasks.
Recently, \citet{li2021benchmarking} and ViTDet~\citep{li2022exploring} employed some upsampling and downsampling modules to adapt the plain ViT for object detection, as shown in Figure~\ref{fig:comparison}(a).
However, under regular training settings
(\ie, apply ImageNet supervised pre-training and fine-tune for 36 epochs), their detection performance is still inferior\footnote{In ViTDet, using regular ImageNet-22K pre-training instead of MAE~\citep{he2021masked} drops 4.0 box AP.} to recent models~\citep{chu2021conditional,dong2021cswin,wang2021pvtv2,wu2022p2t} that well combine image prior. 
Therefore, it is still challenging to design a powerful dense prediction task adapter for ViT.

\begin{figure}[tbp]
    \centering
    \includegraphics[width=0.99\linewidth]{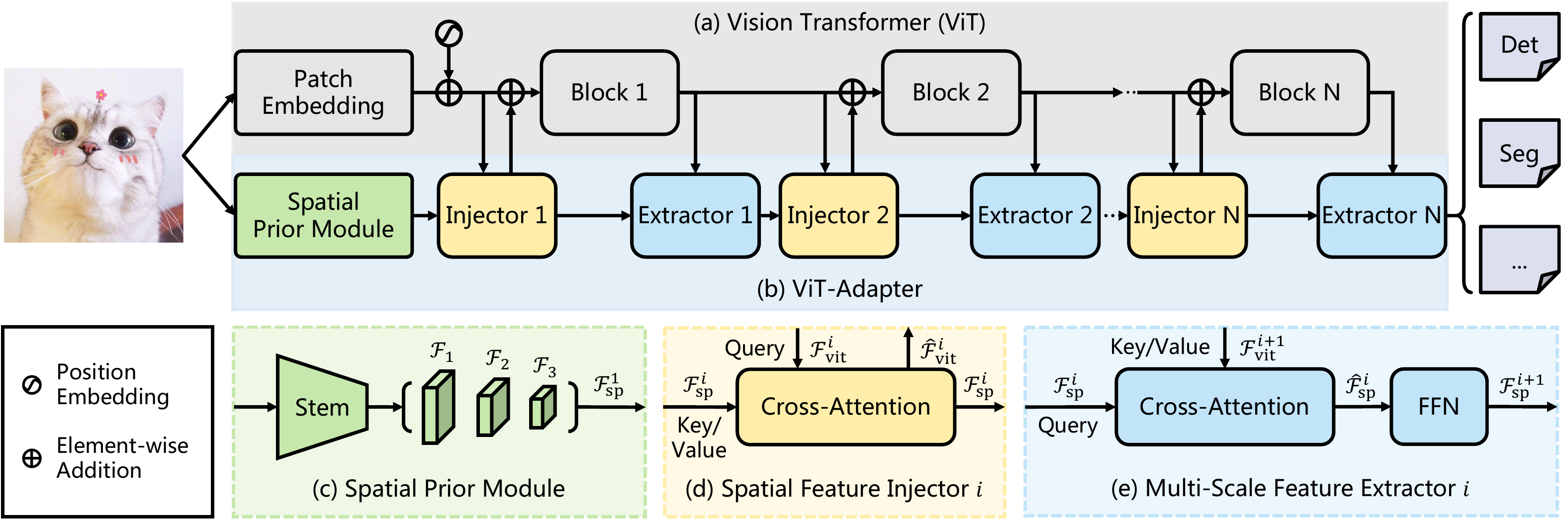}
    \caption{
        \textbf{Overall architecture of ViT-Adapter. }
        (a) The ViT, whose encoder layers are divided into $N$ (usually $N=4$) equal blocks for feature interaction.
        (b) Our ViT-Adapter, which contains three key designs, including
        (c)~a spatial prior module for modeling local spatial contexts from the input image,
        (d)~a spatial feature injector for introducing spatial priors into ViT,
        and (e)~a multi-scale feature extractor for reorganizing multi-scale features from the single-scale features of ViT.
        }
    \label{fig:overall_architecture}
\end{figure}

\section{Vision Transformer Adapter}

\subsection{Overall Architecture}
\label{sec:overall_architecture}

As illustrated in Figure~\ref{fig:overall_architecture}, our model can be divided into two parts. The first part is the plain ViT~\citep{dosovitskiy2020image} that consists of a patch embedding followed by $L$ transformer encoder layers (see Figure~\ref{fig:overall_architecture}(a)).
The second part is the proposed ViT-Adapter as shown in Figure~\ref{fig:overall_architecture}(b), which contains (1) a spatial prior module to capture spatial features from the input image, 
(2) a spatial feature injector to inject spatial priors into the ViT, and (3) a multi-scale feature extractor to extract hierarchical features from the single-scale features of ViT.

For the ViT, the input image is first fed into the patch embedding, where the image is divided into $16\times16$ non-overlapping patches. 
After that, these patches are flattened and projected to $D$-dimensional tokens, and the feature resolution is reduced to 1/16 of the original image.
Then, these tokens added with the position embedding, are passed through $L$ encoder layers.

For the ViT-Adapter, we first feed the input image into the spatial prior module.
$D$-dimensional spatial features of three target resolutions (\ie, 1/8, 1/16, and 1/32) will be collected.
Then, these feature maps are flattened and concatenated as the input for feature interaction.
Specifically, given the number of interactions $N$ (usually $N=4$), we evenly split the transformer encoders of ViT into $N$ blocks, each containing $L/N$ encoder layers.
For the $i$-th block, we first inject spatial priors $\mathcal{F}_\text{sp}^i$ into the block via a spatial feature injector, and then extract hierarchical features from the output of the block by a multi-scale feature extractor.
After $N$ feature interactions, we obtain high-quality multi-scale features, and then we split and reshape the features into three target resolutions 1/8, 1/16, and 1/32.
Finally, we build the 1/4-scale feature map by upsampling the 1/8-scale feature map using a $2\times2$ transposed convolution.
In this way, we obtain a feature pyramid of similar resolutions to ResNet~\citep{he2016deep}, which can be used in various dense prediction tasks.

\subsection{Spatial Prior Module}

Recent studies~\citep{wang2021pvtv2,wu2021cvt,fang2022unleashing,park2022vision} show convolutions can help transformers better capture the local spatial information.
Inspired by this, we introduce the \emph{Spatial Prior Module} (SPM).
It is designed to model the local spatial contexts of images parallel with the patch embedding layer, so as not to alter the original architecture of ViT.

As shown in Figure~\ref{fig:overall_architecture}(c), a standard convolutional stem borrowed from ResNet \citep{he2016deep} is employed, which consists of three convolutions and a max-pooling layer. 
Then, we use a stack of stride-2 3$\times$3 convolutions to double the number of channels and reduce the size of feature maps.
Finally, several 1$\times$1 convolutions are applied at the end to project the feature maps to $D$ dimensions.
In this way, we obtain a feature pyramid $\{\mathcal{F}_1, \mathcal{F}_2, \mathcal{F}_3\}$, which contains $D$-dimensional feature maps with resolutions of 1/8, 1/16, and 1/32. 
Then, we flatten and concatenate these feature maps into feature tokens $\mathcal{F}_{\rm sp}^1\in \mathbb{R}^{(\frac{HW}{8^2}+\frac{HW}{16^2}+\frac{HW}{32^2}) \times D}$, as the input for feature interaction.

\subsection{Feature Interaction}
\label{feature_interaction}
Due to weak prior assumptions, the plain ViT suffers sub-optimal performance on dense prediction tasks compared to vision-specific transformers~\citep{chu2021twins,dong2021cswin,liu2021swin,wang2021pvtv2}.
To alleviate this issue, we propose two feature interaction modules to bridge the feature maps of our SPM and the ViT.
To be specific, the two modules are mainly based on cross-attention, namely \emph{Spatial Feature Injector} and \emph{Multi-Scale Feature Extractor}.

\noindent \textbf{Spatial Feature Injector.}
As shown in Figure~\ref{fig:overall_architecture}(d), this module is used to inject the spatial priors into ViT. Specifically, for the $i$-th block of the ViT, we take the input feature $\mathcal{F}^{i}_{\rm vit}\in\mathbb{R}^{\frac{HW}{16^2} \times D}$ as the query, and the spatial feature $\mathcal{F}^{i}_{\rm sp}\in\mathbb{R}^{(\frac{HW}{8^2}+\frac{HW}{16^2}+\frac{HW}{32^2}) \times D}$ as the key and value. We use cross-attention 
to inject spatial feature $\mathcal{F}^{i}_{\rm sp}$ into the input feature $\mathcal{F}^{i}_{\rm vit}$, 
which can be written as Eqn. \ref{injection}.
\begin{equation}
	\mathcal{\hat{F}}^{i}_{\rm vit}=\mathcal{F}^{i}_{\rm vit} + \gamma^{i} {\rm Attention}({\rm norm}(\mathcal{F}^{i}_{\rm vit}), {\rm norm}(\mathcal{F}^{i}_{\rm sp})),
	\label{injection}
\end{equation}
where the ${\rm norm}(\cdot)$ is LayerNorm~\citep{ba2016layer}, and the attention layer ${\rm Attention}(\cdot)$ suggests using sparse attention. 
In addition, we apply a learnable vector $\gamma^i\in\mathbb{R}^{D}$ to balance the attention layer's output and the input feature $\mathcal{F}^{i}_{\rm vit}$, which is initialized with $\mathbf{0}$. 
This initialization strategy ensures that the feature distribution of $\mathcal{F}^{i}_{\rm vit}$ will not be modified drastically due to the injection of spatial priors, thus making better use of the pre-trained weights of ViT.

\noindent \textbf{Multi-Scale Feature Extractor.}
After injecting the spatial priors into the ViT, we obtain the output feature $\mathcal{F}^{i+1}_{\rm vit}$ by passing $\mathcal{\hat{F}}^{i}_{\rm vit}$ through the encoder layers of the $i$-th block.
Then, we apply a module consisting of a cross-attention layer and a feed-forward network (FFN), to extract multi-scale features, as shown in Figure~\ref{fig:overall_architecture}(e).
This process can be formulated as:
\begin{equation}
	\mathcal{F}^{i+1}_{\rm sp}=\hat{\mathcal{F}}^{i}_{\rm sp} + {\rm FFN}({\rm norm}(\hat{\mathcal{F}}^{i}_{\rm sp})), \\
	\label{cffn}
\end{equation}
\begin{equation}
	\hat{\mathcal{F}}^i_{\rm sp} = \mathcal{F}^{i}_{\rm sp} + {\rm Attention}({\rm norm}(\mathcal{F}^{i}_{\rm sp}), {\rm norm}(\mathcal{F}^{i+1}_{\rm vit})),
	\label{extraction}
\end{equation}
in which we use the spatial feature $\mathcal{F}^{i}_{\rm sp}\in\mathbb{R}^{(\frac{HW}{8^2}+\frac{HW}{16^2}+\frac{HW}{32^2}) \times D}$ as the query, and the output feature $\mathcal{F}^{i+1}_{\rm vit}\in\mathbb{R}^{\frac{HW}{16^2} \times D}$ as the key and value for cross-attention. 
As same as the spatial feature injector, we adopt sparse attention here to reduce computational cost.
The generated spatial feature $\mathcal{F}^{i+1}_{\rm sp}$ will be used as the input of the next spatial feature injector.

\vspace{-0.5em}
\subsection{Architecture Configurations}

We build our ViT-Adapter for 4 different sizes of ViT, including ViT-T, ViT-S, ViT-B, and ViT-L.
For these models, the parameter numbers of our adapters are 2.5M, 5.8M, 14.0M, and 23.7M, respectively.
We employ deformable attention~\citep{zhu2020deformable} as the default sparse attention in our method, where the number of sampling points is fixed to 4, and the number of attention heads is set to 6, 6, 12, and 16.
The number of interactions $N$ is 4, and in the last feature interaction, we stack three multi-scale feature extractors.
Besides, we set the FFN ratio in our adapter to 0.25 to save computational overhead, \ie~the hidden sizes of FFN are 48, 96, 192, and 256 for 4 different adapters.
More details of each configuration are shown in Table~\ref{tab:architecture_configurations} in Appendix~\ref{appendix_config}. 

\vspace{-0.5em}
\section{Experiments}

\begin{table}[t]\small
	\centering
	\renewcommand{\arraystretch}{0.93}
	
    \setlength\tabcolsep{0.12mm}{
    \begin{tabular}{l|c|cccccc|cccccc}
        \toprule
        \multirow{2}{*}{Method} & \#Param & 
        \multicolumn{6}{c|}{Mask R-CNN 1$\times$ schedule} & \multicolumn{6}{c}{Mask R-CNN 3$\times$+MS schedule}\\
        ~ & (M)
        & $\rm AP^b$ & $\rm AP^b_{50}$ & $\rm AP^b_{75}$ & $\rm AP^m$ & $\rm AP^m_{50}$ & $\rm AP^m_{75}$ 
        & $\rm AP^b$ & $\rm AP^b_{50}$ & $\rm AP^b_{75}$ & $\rm AP^m$ & $\rm AP^m_{50}$ & $\rm AP^m_{75}$   \\
	    \midrule
        PVT-Tiny~\citep{wang2021pyramid} & 32.9 
        & 36.7 & 59.2 & 39.3 & 35.1 & 56.7 & 37.3 & 39.8 & 62.2 & 43.0 & 37.4 & 59.3 & 39.9  \\
        PVTv2-B1 \citep{wang2021pvtv2} & 33.7 
        & 41.8 & 64.3 & 45.9 & 38.8 & 61.2 & 41.6 & 44.9 & 67.3 & 49.4 & 40.8 & 64.0 & 43.8  \\
        ViT-T~\citep{li2021benchmarking} & 26.1
        & 35.5 & 58.1 & 37.8 & 33.5 & 54.9 & 35.1 & 40.2 & 62.9 & 43.5 & 37.0 & 59.6 & 39.0  \\
        ViTDet-T~\citep{li2022exploring} & 26.6 & 35.7 & 57.7 & 38.4 & 33.5 & 54.7 & 35.2 &
        40.4 & 63.3 & 43.9 & 37.1 & 60.1 & 39.3\\
        \rowcolor{gray!20} 
        ViT-Adapter-T (ours) & 28.1 
        & 41.1 & 62.5 & 44.3 & 37.5 & 59.7 & 39.9 & 46.0 & 67.6 & 50.4 & 41.0 & 64.4 & 44.1  \\
        \midrule
        PVT-Small~\citep{wang2021pyramid} & 44.1
        & 40.4 & 62.9 & 43.8 & 37.8 & 60.1 & 40.3 & 43.0 & 65.3 & 46.9 & 39.9 & 62.5 & 42.8  \\
        PVTv2-B2~\citep{wang2021pvtv2} & 45.0 
        & 45.3 & 67.1 & 49.6 & 41.2 & 64.2 & 44.4 & 47.8 & 69.7 & 52.6 & 43.1 & 66.8 & 46.7  \\
        Swin-T~\citep{liu2021swin} & 47.8 
        & 42.7 & 65.2 & 46.8 & 39.3 & 62.2 & 42.2 & 46.0 & 68.1 & 50.3 & 41.6 & 65.1 & 44.9  \\
        ConvNeXt-T~\citep{liu2022convnet} & 48.1 
        & 44.2 & 66.6 & 48.3 & 40.1 & 63.3 & 42.8 & 46.2 & 67.9 & 50.8 & 41.7 & 65.0 & 44.9  \\
        Focal-T~\citep{yang2021focal} & 48.8 
        & 44.8 & 67.7 & 49.2 & 41.0 & 64.7 & 44.2 & 47.2 & 69.4 & 51.9 & 42.7 & 66.5 & 45.9  \\
        ViT-S~\citep{li2021benchmarking} & 43.8 
        & 40.2 & 63.1 & 43.4 & 37.1 & 59.9 & 39.3 & 44.0 & 66.9 & 47.8 & 39.9 & 63.4 & 42.2  \\
        ViTDet-S~\citep{li2022exploring} & 45.7 & 40.6 & 63.3 & 43.5 & 37.1 & 60.0 & 38.8 & 44.5 & 66.9 & 48.4 & 40.1 & 63.6 & 42.5 \\
        \rowcolor{gray!20} 
        ViT-Adapter-S (ours)  & 47.8 
        & 44.7 & 65.8 & 48.3 & 39.9 & 62.5 & 42.8 & 48.2 & 69.7 & 52.5 & 42.8 & 66.4 & 45.9  \\
        \midrule
        PVTv2-B5~\citep{wang2021pvtv2} & 101.6 
        & 47.4 & 68.6 & 51.9 & 42.5 & 65.7 & 46.0 & 48.4 & 69.2 & 52.9 & 42.9 & 66.6 & 46.2  \\
        Swin-B~\citep{liu2021swin} & 107.1 
        & 46.9 & - & - & 42.3 & - & - & 48.6 & 70.0 & 53.4 & 43.3 & 67.1 & 46.7  \\
        ViT-B~\citep{li2021benchmarking} & 113.6 
        & 42.9 & 65.7 & 46.8 & 39.4 & 62.6 & 42.0 & 45.8 & 68.2 & 50.1 & 41.3 & 65.1 & 44.4  \\
        ViTDet-B~\citep{li2022exploring} & 121.3 & 43.2 & 65.8 & 46.9 & 39.2 & 62.7 & 41.4
        & 46.3 & 68.6 & 50.5 & 41.6 & 65.3 & 44.5\\
        \rowcolor{gray!20} 
        ViT-Adapter-B (ours) & 120.2 
        & 47.0 & 68.2 & 51.4 & 41.8 & 65.1 & 44.9 & 49.6 & 70.6 & 54.0 & 43.6 & 67.7 & 46.9  \\
	    \midrule
        ViT-L$^\dagger$~\citep{li2021benchmarking} & 337.3 
        & 45.7 & 68.9 & 49.4 & 41.5 & 65.6 & 44.6 & 48.3 & 70.4 & 52.9 & 43.4 & 67.9 & 46.6  \\
        ViTDet-L$^\dagger$~\citep{li2022exploring} & 350.9 & 46.2 & 69.2 & 50.3 & 41.4 & 65.8 & 44.1 & 49.1 & 71.5 & 53.8 & 44.0 & 68.5 & 47.6\\
        \rowcolor{gray!20} 
        ViT-Adapter-L$^\dagger$ (ours) & 347.9
        & 48.7 & 70.1 & 53.2 & 43.3 & 67.0 & 46.9 & 52.1 & 73.8 & 56.5 & 46.0 & 70.5 & 49.7  \\
        \bottomrule
    \end{tabular}}
    
    \caption{\textbf{Object detection and instance segmentation with Mask R-CNN on COCO val2017.} 
    For fair comparison, we initialize all ViT-T/S/B models with the regular ImageNet-1K pre-training~\citep{touvron2021training}, and ViT-L$^\dagger$ with the ImageNet-22K weights from~\citep{steiner2021train}. 
	}
    \label{tab:results_detection_mask}
\end{table}

Previous work~\citep{wang2021pyramid} has shown that the pyramid prior is beneficial to dense prediction, but brings little gains to image classification.
Therefore, in this study, we focus on how to better adapt readily available pre-trained ViTs to dense prediction tasks.
We hope this method will also help decouple the model design of upstream pre-training and downstream fine-tuning.

\subsection{Object Detection And Instance Segmentation}
\label{sec:detection_settings}

\noindent \textbf{Settings.}
Our detection experiments are based on MMDetection \citep{mmdetection} and the COCO \citep{lin2014microsoft} dataset.
We use 4 mainstream detectors to evaluate our ViT-Adapter, including Mask R-CNN \citep{he2017mask}, Cascade Mask R-CNN \citep{cai2019cascade}, ATSS \citep{zhang2020bridging}, and GFL \citep{li2020generalized}. 
To save time and memory, we refer to \citep{li2021benchmarking} and modify the $L$-layer ViT to use 14$\times$14 window attention except for layers spaced at an interval of $L/4$.
Following common practices~\citep{wang2021pyramid}, we adopt 1$\times$ or 3$\times$ training schedule (\ie, 12 or 36 epochs) with a batch size of 16, and AdamW~\citep{loshchilov2017decoupled} optimizer with an initial learning rate of $1\times10^{-4}$ and a weight decay of 0.05. 

\begin{table}[t]\small
    \renewcommand{\arraystretch}{0.95}
    
 \begin{minipage}{0.35\textwidth}%
    \centering
	{
	\setlength\tabcolsep{0.0mm}{
    \begin{tabular}{lcccc}
        \toprule
        Method & $\rm AP^b$ & $\rm AP^b_{50}$ & $\rm AP^b_{75}$ & \#P   \\
	    \midrule
	    \multicolumn{5}{c}{Cascade Mask R-CNN 3$\times$+MS schedule} \\
        Swin-T \citep{liu2021swin} & 50.5 & 69.3 & 54.9 & 86M  \\ 
        Shuffle-T \citep{huang2021shuffle} & 50.8 & 69.6 & 55.1 & 86M \\
        PVTv2-B2 \citep{wang2021pvtv2} & 51.1 & 69.8 & 55.3 & 83M \\
        Focal-T \citep{yang2021focal}  & 51.5 & 70.6 & 55.9 & 87M  \\
        ViT-S~\citep{li2021benchmarking} & 47.9 & 67.1 & 51.7 & 82M   \\
        \rowcolor{gray!20}
        ViT-Adapter-S (ours) & 51.5 & 70.1 & 55.8 & 86M  \\
	    \midrule
	    Swin-B \citep{liu2021swin} & 51.9 & 70.9 & 57.0 & 145M  \\ 
	    Shuffle-B \citep{huang2021shuffle} & 52.2 & 71.3 & 57.0 & 145M  \\ 
	    ViT-B \citep{li2021benchmarking} & 50.1 & 69.3 & 54.3 & 151M \\ 
	    \rowcolor{gray!20}
	    ViT-Adapter-B (ours) & 52.1 & 70.6 & 56.5 &  158M \\ 
	    \bottomrule
    \end{tabular}}
	}
 \end{minipage}
 \hspace{6.7em}
 \begin{minipage}{0.4\textwidth}%
	\centering
	{
	\setlength\tabcolsep{0.mm}{
    \begin{tabular}{lcccc}
        \toprule
        Method & $\rm AP^b$ & $\rm AP^b_{50}$ & $\rm AP^b_{75}$ & \#P  \\
	    \midrule
	    \multicolumn{5}{c}{ATSS 3$\times$+MS schedule} \\
	    Swin-T \citep{liu2021swin} & 47.2 & 66.5 & 51.3 & 36M  \\ 
	    Focal-T \citep{yang2021focal} & 49.5 & 68.8 & 53.9 & 37M \\
	    PVTv2-B2 \citep{wang2021pvtv2} & 49.9 & 69.1 & 54.1 & 33M  \\
	   ViT-S~\citep{li2021benchmarking} & 45.2 & 64.8 & 49.0 & 32M   \\ 
	   \rowcolor{gray!20}
	    ViT-Adapter-S (ours) & 49.6 & 68.5 & 54.0 & 36M  \\ 
	    \midrule
	    \multicolumn{5}{c}{GFL 3$\times$+MS schedule} \\
	    Swin-T \citep{liu2021swin} & 47.6 & 66.8 & 51.7 & 36M  \\ 
	    PVTv2-B2 \citep{wang2021pvtv2} & 50.2 & 69.4 & 54.7 & 33M  \\
	    ViT-S~\citep{li2021benchmarking} & 46.0 & 65.5 & 49.7 & 32M   \\
	    \rowcolor{gray!20}
	    ViT-Adapter-S (ours) & 50.0 & 69.1 & 54.3 & 36M  \\
	    \bottomrule
    \end{tabular}}
    }
\end{minipage}
\caption{\textbf{Object detection with different frameworks on COCO val2017.} 
    For fair comparison, we initialize all ViT-S/B models with the regular ImageNet-1K pre-training \citep{touvron2021training}.
    ``\#P" denotes the number of parameters. 
    ``MS" means multi-scale training.
    }
\label{tab:results_detection_strong_heads}
\end{table}

\noindent \textbf{Results with ImageNet-1K Pre-training.} 
In Table \ref{tab:results_detection_mask} and Table \ref{tab:results_detection_strong_heads}, we apply the DeiT~\citep{touvron2021training} released ImageNet-1K weights (without distillation) as the initialization for all ViT-T/S/B models. 
We compare our ViT-Adapter with two related approaches~\citep{li2021benchmarking,li2022exploring} and multiple representative vision-specific backbones~\citep{wang2021pyramid,wang2021pvtv2,huang2021shuffle,liu2021swin,yang2021focal}. 
As we can see, when using regular training settings for fair comparison, the detection performance of ViT~\citep{li2021benchmarking} and ViTDet~\citep{li2022exploring} is inferior to recent vision-specific models.
For example, with Mask R-CNN and 3$\times$+MS schedule, ViT-S and ViTDet-S are 3.8 AP$^{\rm b}$ and 3.3 AP$^{\rm b}$ lower than PVTv2-B2~\citep{wang2021pvtv2} respectively.
Differently, our ViT-Adapter-S outperforms these two approaches by clear margins and even 0.4 AP$^{\rm b}$ higher than PVTv2-B2.
This observation can also be seen in the experiments of three other detectors, including Cascade Mask R-CNN, ATSS, and GFL.
These results indicate that, \emph{with only the regular ImageNet-1K pre-training}, ViT-Adapter can promote the plain ViT to attain similar or even superior performance than these vision-specific transformers.

\noindent \textbf{Results with ImageNet-22K Pre-training.} 
In Table \ref{tab:results_detection_mask}, we employ the ImageNet-22K pre-trained weights from AugReg~\citep{steiner2021train} to initialize all ViT-L models, including ViT~\citep{li2021benchmarking},  ViTDet~\citep{li2022exploring}, and our ViT-Adapter. 
It can be seen that, when training Mask R-CNN with 3$\times$+MS schedule, our ViT-Adapter-L$^\dagger$ brings 3.8 AP$^{\rm b}$ and 3.0 AP$^{\rm b}$ improvements over ViT-L$^\dagger$~\citep{li2021benchmarking} and ViTDet-L$^\dagger$~\citep{li2022exploring}, respectively.

\begin{table}[t]\small
    \centering
    \renewcommand{\arraystretch}{0.93}
	\setlength\tabcolsep{1.41mm}{
	\begin{tabular}{l|c|c|ccc|ccc}
		\toprule
	    \multirow{2}{*}{Method} &   \multirow{2}{*}{Pre-train}  &   \multirow{2}{*}{Crop Size} & \multicolumn{3}{c|}{Semantic FPN 80k}   & \multicolumn{3}{c}{UperNet 160k}  \\
	    & & & \#Param & mIoU & +MS  & \#Param & mIoU & +MS \\
		\midrule
        PVT-Tiny \citep{wang2021pyramid} & IN-1K & 512$\times$512 & 17.0M & 36.6 & 37.3 & 43.2M & 38.5 & 39.0 \\
        ViT-T~\citep{li2021benchmarking} & IN-1K & 512$\times$512 & 10.2M & 39.4 & 40.5 & 34.1M & 41.7 & 42.6 \\
        \rowcolor{gray!20}
        ViT-Adapter-T (ours) & IN-1K & 512$\times$512 & 12.2M & 41.7 & 42.1 & 36.1M & 42.6 & 43.6 \\
        
	    \midrule
	    PVT-Small \citep{wang2021pyramid} & IN-1K & 512$\times$512 & 28.2M & 41.9 & 42.3 & 54.5M & 43.7 & 44.0\\
		PVTv2-B2 \citep{wang2021pvtv2} & IN-1K & 512$\times$512 & 29.1M & 45.2 & 45.7 & - & - & -\\
		Swin-T \citep{liu2021swin} & IN-1K & 512$\times$512 & 31.9M & 41.5 & - & 59.9M & 44.5 & 45.8  \\
		Twins-SVT-S \citep{chu2021twins} & IN-1K & 512$\times$512 & 28.3M & 43.2 & - & 54.4M & 46.2 & 47.1\\
		
        ViT-S~\citep{li2021benchmarking} & IN-1K & 512$\times$512 & 27.8M & 44.6 & 45.8 & 53.6M & 44.6 & 45.7 \\
        \rowcolor{gray!20}
        ViT-Adapter-S (ours) & IN-1K & 512$\times$512 & 31.9M & 46.1 & 46.6 & 57.6M & 46.2 & 47.1 \\
        
		\midrule
		Swin-B \citep{liu2021swin} & IN-1K & 512$\times$512 & 91.2M & 46.0 & - & 121.0M & 48.1 & 49.7  \\
		Twins-SVT-L \citep{chu2021twins} & IN-1K & 512$\times$512 & 103.7M & 46.7 & - & 133.0M & 48.8 & 50.2  \\
        ViT-B~\citep{li2021benchmarking} & IN-1K & 512$\times$512 & 98.0M & 46.4 & 47.6 & 127.3M & 46.1 & 47.1 \\
        \rowcolor{gray!20}
        ViT-Adapter-B (ours) & IN-1K & 512$\times$512 & 104.6M & 47.9 & 48.9 & 133.9M & 48.8 & 49.7 \\		
        \midrule
        Swin-B$^\dagger$ \citep{liu2021swin} & IN-22K & 640$\times$640 & - & - & - & 121.0M & 50.0 & 51.7  \\
        Swin-L$^\dagger$ \citep{liu2021swin} & IN-22K & 640$\times$640 & - & - & - & 234.0M & 52.1 & 53.5  \\
        \rowcolor{gray!20}
        ViT-Adapter-B$^\dagger$ (ours) & IN-22K & 512$\times$512 & 104.6M & 50.7 & 51.9 & 133.9M & 51.9 & 52.5 \\
        \rowcolor{gray!20}
        ViT-Adapter-L$^\dagger$ (ours) & IN-22K & 512$\times$512 & 332.0M & 52.9 & 53.7 & 363.8M & 53.4 & 54.4 \\
        \rowcolor{yellow!15}
        ViT-Adapter-L$^\bigstar$ (ours) & MM & 512$\times$512 & 332.0M & 54.2  & 54.7 & 363.8M & 55.0 & 55.4 \\
        
		\bottomrule
	\end{tabular}}
	\caption{\textbf{Semantic segmentation on the ADE20K val.}
	Semantic FPN~\citep{kirillov2019panoptic} and UperNet~\citep{xiao2018unified} are used as segmentation frameworks.
	``IN-1K/22K'' and ``MM'' represent ImageNet-1K/22K and multi-modal pre-training, respectively.
	``MS" means multi-scale testing.}
	\label{tab:result_ade}
\end{table}

\begin{wraptable}{r}{6.9cm}\small
\renewcommand\arraystretch{0.95}
\centering
\vspace{-1.2em}
\setlength{\tabcolsep}{0.5mm}{
    \begin{tabular}[t]{c|c|cc}
    \toprule
    Method & Pre-train & $\rm AP^b$ &$\rm AP^m$   \\
    \midrule
    \multirow{3}{*}{ \shortstack{Swin-B \\(Mask R-CNN 3$\times$+MS)}} & ImageNet-1K & 48.6 & 43.3 \\
      & ImageNet-22K & 49.6 & 44.3 \\
      & Multi-Modal\cellcolor[HTML]{FFFADF} & N/A\cellcolor[HTML]{FFFADF} & N/A\cellcolor[HTML]{FFFADF} \\
    \midrule
    \multirow{3}{*}{ \shortstack{ViT-Adapter-B \\(Mask R-CNN 3$\times$+MS)}} & ImageNet-1K & 49.6 & 43.6 \\
                                   & ImageNet-22K & 50.5 & 44.6\\
                                   & Multi-Modal\cellcolor[HTML]{FFFADF} & \textbf{51.2}\cellcolor[HTML]{FFFADF} & \textbf{45.3}\cellcolor[HTML]{FFFADF}  \\
    \bottomrule
    \end{tabular}}
    \caption{\textbf{Comparison of different pre-trained weights.}
    Our method retains the flexibility of ViT and thus could benefit from advanced multi-modal pre-training~\citep{zhu2021uni}.
}
\vspace{-1em}
\label{tab:det_pretrain}
\end{wraptable}

\noindent \textbf{Results with Multi-Modal Pre-training.} 
In this experiment, we study the effect of multi-modal pre-training. 
Specifically, we fine-tune the ViT-Adapter-B with Mask R-CNN for the 3$\times$+MS schedule using different pre-trained weights.
As shown in Table~\ref{tab:det_pretrain}, simply replacing the ImageNet-22K pre-training~\citep{steiner2021train} with the multi-modal pre-training~\citep{zhu2021uni} gives us a significant gain of 0.7 AP$^\text{b}$ and AP$^\text{m}$.
These results indicate that our method can easily derive considerable benefits from advanced multi-modal pre-training, which is difficult for vision-specific models like Swin.

\vspace{-1em}
\subsection{Semantic Segmentation}

\noindent \textbf{Settings.}
We evaluate our ViT-Adapter on semantic segmentation with the ADE20K~\citep{zhou2017scene} dataset and MMSegmentation~\citep{mmseg2020} codebase.
Both Semantic FPN~\citep{kirillov2019panoptic} and UperNet~\citep{xiao2018unified} are employed as the basic frameworks.
For Semantic FPN, we apply the settings of PVT~\citep{wang2021pyramid} and train the models for 80k iterations. 
For UperNet, we follow the settings of Swin~\citep{liu2021swin} to train it for 160k iterations. 

\vspace{-1em}
\noindent \textbf{Results with ImageNet-1K Pre-training.} 
In Table \ref{tab:result_ade}, we report the semantic segmentation results in terms of single-scale and multi-scale (MS) mIoU.
As same as Section~\ref{sec:detection_settings}, we initialize all ViT-T/S/B models with the DeiT~\citep{touvron2021training} released ImageNet-1K weights.
It shows that, under comparable model sizes, our method surpasses the ViT~\citep{li2021benchmarking} and many representative vision-specific transformers~\citep{wang2021pyramid,wang2021pvtv2,liu2021swin,chu2021twins}.
For instance, our ViT-Adapter-S achieves 47.1 MS mIoU with UperNet, outperforming many strong counterparts such as Swin-T.
Similarly, ViT-Adapter-B reports a competitive performance of 49.7 MS mIoU, which is 2.6 points higher than ViT-B and on par with Swin-B and Twins-SVT-L.
These fair comparisons using only regular ImageNet-1K pre-training~\citep{touvron2021training} demonstrate the effectiveness and universality of our ViT-Adapter.

\noindent \textbf{Results with ImageNet-22K Pre-training.} 
When using the ImageNet-22K pre-trained weights \citep{steiner2021train}, our ViT-Adapter-B$^\dagger$ attains 51.9 mIoU and 52.5 MS mIoU with UperNet, exceeding Swin-B$^\dagger$ by at least 0.8 mIoU.
Similarly, 
ViT-Adapter-L$^\dagger$ yields the results of 53.4 mIoU and 54.4 MS mIoU, which is outstanding from the counterparts like Swin-L$^\dagger$.
These significant and consistent improvements over different model sizes suggest that our method can cover the shortage of plain ViT, making it more suitable for semantic segmentation.

\noindent \textbf{Results with Multi-Modal Pre-training.} 
Here, we apply the multi-modal pre-trained weights from Uni-Perceiver~\citep{zhu2021uni} for semantic segmentation. 
As shown in Table~\ref{tab:result_ade}, for Semantic FPN and UperNet, replacing the ImageNet-22K pre-training with multi-modal pre-training benefits our ViT-Adapter-L$^\bigstar$ with impressive gains of 1.3 mIoU and 1.6 mIoU, respectively.


\vspace{-0.5em}
\subsection{Comparisons with State-of-the-Arts}

\noindent\textbf{Settings.}
We conduct experiments to combine our ViT-Adapter with state-of-the-art detection/segmentation frameworks, including HTC++~\citep{liu2021swin} (without extra detection dataset) and Mask2Former~\citep{cheng2021masked}, and recent multi-modal pre-training BEiTv2~\citep{peng2022beitv2}. 
The experimental settings are listed in Appendix~\ref{sec:appendix_detection} and \ref{sec:appendix_segmentation}.

\begin{wraptable}{r}{7.4cm}\small
\renewcommand\arraystretch{0.95}
\vspace{-1.em}
{
\begin{minipage}{0.1\textwidth}
\centering
{\setlength{\tabcolsep}{0.1mm}{
    \begin{tabular}[t]{l|cc}
    \toprule
    COCO test-dev &  AP$\rm^{b}$ & AP$\rm^{m}$   \\
    \midrule
     CB-Swin-L & 60.1 & 52.3 \\
     SwinV2-L & 60.8 & 52.7 \\
     \textbf{ViT-Adapter-L}\cellcolor[HTML]{FFFADF} &  \textbf{60.9}\cellcolor[HTML]{FFFADF} & \textbf{53.0}\cellcolor[HTML]{FFFADF} \\
    \bottomrule
    \end{tabular}}
}
\end{minipage}
\hspace{1em}
\begin{minipage}{0.53\textwidth}
\centering
{
\setlength{\tabcolsep}{0.mm}{
    \begin{tabular}[t]{l|c}
    \toprule
    ADE20K val &  mIoU  \\
    \midrule
     FD-SwinV2-G & 61.4 \\
     \textbf{ViT-Adapter-L}\cellcolor[HTML]{FFFADF} & \textbf{61.5}\cellcolor[HTML]{FFFADF} \\
     BEiT3(w/ ViT-Adapter) \cellcolor[HTML]{FFFADF}&  \textbf{62.8}\cellcolor[HTML]{FFFADF} \\
    \bottomrule
    \end{tabular}}
}
\end{minipage}
}
\vspace{-1em}
\caption{\textbf{Comparison with previous SOTA.}}
\label{tab:small_sota}
\end{wraptable}
\noindent\textbf{Results.}
As shown in Table~\ref{tab:small_sota}, our method reaches state-of-the-art performance.
While these results may be partly due to the effectiveness of advanced pre-training, our study demonstrates that plain backbone detectors/segmenters can challenge the
entrenched position of hierarchical backbones.

\subsection{Ablation Study}
\label{sec:ablation}
\begin{figure}[t]
    \centering
    \includegraphics[width=0.9\linewidth]{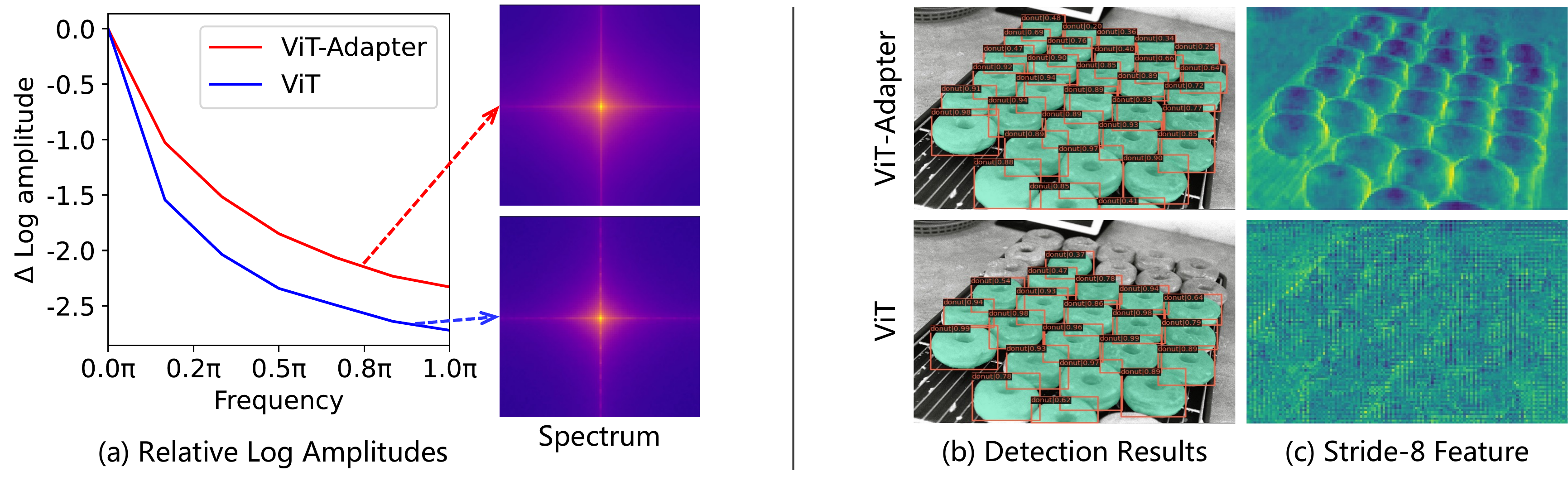}
    \vspace{-0.5em}
    \caption{
        \textbf{ViT \emph{vs.} ViT-Adapter Feature.}
        (a) Relative log amplitudes of Fourier transformed feature maps.
        (b) Detection results.
        (c) Stride-8 feature map. 
        Compared to the ViT baseline \citep{li2021benchmarking}, our ViT-Adapter captures more high-frequency signals, and produces more fine-grained features with rich edges and textures, which is of great help for dense prediction.
        }
    \label{fig:feature_map}
\end{figure}

\noindent\textbf{ViT \emph{vs.} ViT-Adapter Feature.}~Recent works \citep{park2022vision,si2022inception} show that ViT presents the characteristics of learning low-frequency global signals, while CNN tends to extract high-frequency information (\eg, local edges and textures).
To show the difference between the features of ViT and ViT-Adapter,
we first use Fourier analysis as a toolkit for visualization.
As shown in Figure \ref{fig:feature_map}(a), the Fourier spectrum and relative log amplitudes of the Fourier transformed feature maps (average over 100 images) indicate that ViT-Adapter captures more high-frequency signals than the ViT~\citep{li2021benchmarking} baseline.
In addition, we also visualize the stride-8 feature map in Figure~\ref{fig:feature_map} (b)(c), which shows that the features of ViT are blurry and coarse.
In contrast, our features are more fine-grained and have more local edges and textures.
This observation demonstrates that our method grafts the merit of CNN for capturing high-frequency information to ViT.

\begin{table}[t]\small
	\renewcommand{\arraystretch}{0.95}
	
	\begin{minipage}{0.35\textwidth}%
	\centering
    \setlength\tabcolsep{0.5mm}
    \begin{tabular}{l|ccc|c|ccc}
        \toprule
        \multirow{2}{*}{Method} & \multicolumn{3}{c|}{Components} & Interaction & \multicolumn{3}{c}{Mask R-CNN 1$\times$}\\
          & SPM & Injector & Extractor & Mode & ${\rm AP^b}$ & ${\rm AP^m}$ & \#Param \\
	    \midrule
        ViT-S~\citep{li2021benchmarking}  & ~ & ~ & ~  & - & 40.2 & 37.1 & 43.8M  \\ 
        Variant 1 & \checkmark  & ~  &  ~ & Add & 41.6 & 38.0 & 45.1M  \\ 
        Variant 2 & \checkmark  & \checkmark & ~ & Attention & 42.6 & 38.8 & 46.6M  \\ 
        \rowcolor{gray!20}
        ViT-Adapter-S (ours) & \checkmark  & \checkmark   & \checkmark & Attention & \textbf{44.7} & \textbf{39.9} & 47.8M   \\ 
	    \bottomrule
    \end{tabular}
    \end{minipage}
    \hspace{17.65em}
    \begin{minipage}{0.95\textwidth}%
    \setlength\tabcolsep{0.9mm}
    \begin{tabular}{c|ccc}
        \toprule
        $N$ & ${\rm AP^b}$ & ${\rm AP^m}$ & \#Param \\
	    \midrule
         0 & 40.2 & 37.1 & 43.8M \\ 
         1 & 43.2 & 38.9 & 45.5M \\ 
         2 & 43.9 & 39.4 & 46.2M \\ 
         \rowcolor{gray!20}
         4 & \textbf{44.7} & \textbf{39.9} & 47.8M \\ 
         6 & 44.7 & 39.8 & 49.4M \\ 
	    \bottomrule
    \end{tabular}
    \end{minipage}
    \caption{\textbf{Ablation studies of ViT-Adapter.}
	(Left) ablation of key components. 
	Our proposed components collectively bring 4.5 AP$\rm ^{b}$ and 2.8 AP$\rm ^{m}$ gains.
	(Right) ablation of the number of interactions $N$. The model gives the best performance when $N=4$.
	SPM is short for the spatial prior module.
	}
    \label{tab:ablation_component}
\end{table}
\noindent \textbf{Ablation for Components.}
To investigate the contribution of each key design, we gradually extend the ViT-S baseline~\citep{li2021benchmarking} to our ViT-Adapter-S. 
All models are trained with Mask R-CNN for 1$\times$ schedule.
As shown in the left side of Table \ref{tab:ablation_component}, by directly resizing and adding the spatial features from SPM, our variant 1 improves 1.4 AP$^\text{b}$ and 0.9 AP$^\text{m}$ over the baseline, showing that local spatial information is essential for dense prediction.
From variant 2, we find that the spatial feature injector further boosts the performance by 1.0 AP$^\text{b}$ and 0.8 AP$^\text{m}$.
This observation illustrates that cross-attention is a more flexible way to inject spatial features.
Moreover, we employ the multi-scale feature extractor to reconstruct hierarchical features, which brings 2.1 AP$^\text{b}$ and 1.1 AP$^\text{m}$ gains, alleviating ViT's drawback of single-scale features.
In summary, our proposed components are each necessary and collectively create 4.5 AP$^\text{b}$ and 2.8 AP$^\text{m}$ improvements.

\noindent \textbf{Number of Interactions.}
In the right side of Table \ref{tab:ablation_component}, we study the effect of the number of interactions. 
Specifically, we build several ViT-Adapter-S variants with different numbers of interactions.
We observe that the model accuracy saturates when $N$ goes larger, and applying more interactions cannot monotonically promote the performance.
Therefore, we empirically set $N$ to 4 by default. 

\begin{table}[t!]\small
\centering
\renewcommand\arraystretch{0.95}
\setlength{\tabcolsep}{1.02mm}{
    \begin{tabular}[t]{lc|cc|ccccc}
    \toprule
    Attention Mechanism & Complexity &$\rm AP^b$ & $\rm AP^m$ &FLOPs & \#Param & Train Time & Memory  \\
    \midrule
    Global Attention~\citep{vaswani2017attention} & Quadratic & 43.7 & 39.3 & 1080G  & 50.3M & 1.61s & *19.0G  \\
    CSwin Attention~\citep{dong2021cswin} & Linear & 43.5 & 39.2 & 456G & 50.3M & 0.56s & 15.6G\\
    Pale Attention~\citep{wu2022pale} & Linear &44.2 & 39.8  & 458G & 50.3M & 0.75s & 17.4G\\
    \rowcolor{gray!20}
    Deformable Attention~\citep{zhu2020deformable} & Linear & \textbf{44.7} & \textbf{39.9}  & \textbf{403G} & \textbf{47.8M} & \textbf{0.36s} & \textbf{13.7G} \\
    \bottomrule
    \end{tabular}}
\caption{\textbf{Ablation of using different attention mechanisms in our adapter.}
The per-iteration training time and GPU training memory are measured by A100 GPUs with per-GPU batch size 2 and FP16 training.
``*" indicates using activation checkpointing to save training memory.
}
\label{tab:ablation_attn}
\end{table}

\noindent \textbf{Attention Type.}
Our method is a general framework in which the attention mechanism is replaceable.
To verify this, we adopt ViT-Adapter-S as the basic model and study 4 different attention mechanisms.
As shown in Table~\ref{tab:ablation_attn}, sparse attention with linear complexity is more suitable for our adapter than global attention with quadratic complexity. 
We ended up using deformable attention~\citep{zhu2020deformable} as the default configuration.
Notably, it can be replaced by other more advanced attention mechanisms in the future to further boost performance.

\section{Conclusion}
This work explores a new paradigm, namely ViT-Adapter, to bridge the performance gap between the plain ViT and vision-specific transformers on dense prediction tasks.
Without modifying the inherent architecture, we flexibly inject image-related inductive biases into the ViT and reconstruct fine-grained multi-scale features required by dense predictions.
Extensive experiments on object detection, instance segmentation, and semantic segmentation show that our method can achieve comparable or even better performance than well-designed vision-specific transformers, and further derive considerable benefits from advanced multi-modal pre-training.

\section*{Acknowledgement}
This work is partly supported by the National Natural Science Foundation of China (Grant No. 61672273, 61832008), and the Shanghai Committee of Science and Technology (Grant No. 21DZ1100100).

\bibliography{iclr2023_conference}

\begin{thebibliography}{80}
\providecommand{\natexlab}[1]{#1}
\providecommand{\url}[1]{\texttt{#1}}
\expandafter\ifx\csname urlstyle\endcsname\relax
  \providecommand{\doi}[1]{doi: #1}\else
  \providecommand{\doi}{doi: \begingroup \urlstyle{rm}\Url}\fi

\bibitem[Ba et~al.(2016)Ba, Kiros, and Hinton]{ba2016layer}
Jimmy~Lei Ba, Jamie~Ryan Kiros, and Geoffrey~E Hinton.
\newblock Layer normalization.
\newblock \emph{arXiv preprint arXiv:1607.06450}, 2016.

\bibitem[Bahng et~al.(2022)Bahng, Jahanian, Sankaranarayanan, and
  Isola]{bahng2022exploring}
Hyojin Bahng, Ali Jahanian, Swami Sankaranarayanan, and Phillip Isola.
\newblock Exploring visual prompts for adapting large-scale models.
\newblock \emph{arXiv preprint arXiv:2203.17274}, 1\penalty0 (3):\penalty0 4,
  2022.

\bibitem[Bao et~al.(2022)Bao, Dong, and Wei]{bao2021beit}
Hangbo Bao, Li~Dong, and Furu Wei.
\newblock Beit: Bert pre-training of image transformers.
\newblock In \emph{ICLR}, 2022.

\bibitem[Bodla et~al.(2017)Bodla, Singh, Chellappa, and Davis]{bodla2017soft}
Navaneeth Bodla, Bharat Singh, Rama Chellappa, and Larry~S Davis.
\newblock Soft-nms--improving object detection with one line of code.
\newblock In \emph{ICCV}, pp.\  5561--5569, 2017.

\bibitem[Bolya et~al.(2020)Bolya, Foley, Hays, and Hoffman]{bolya2020tide}
Daniel Bolya, Sean Foley, James Hays, and Judy Hoffman.
\newblock Tide: A general toolbox for identifying object detection errors.
\newblock In \emph{ECCV}, pp.\  558--573, 2020.

\bibitem[Caesar et~al.(2018)Caesar, Uijlings, and Ferrari]{caesar2018coco}
Holger Caesar, Jasper Uijlings, and Vittorio Ferrari.
\newblock Coco-stuff: Thing and stuff classes in context.
\newblock In \emph{CVPR}, pp.\  1209--1218, 2018.

\bibitem[Cai \& Vasconcelos(2019)Cai and Vasconcelos]{cai2019cascade}
Zhaowei Cai and Nuno Vasconcelos.
\newblock Cascade r-cnn: high quality object detection and instance
  segmentation.
\newblock \emph{TPAMI}, 43\penalty0 (5):\penalty0 1483--1498, 2019.

\bibitem[Chen et~al.(2019{\natexlab{a}})Chen, Pang, Wang, Xiong, Li, Sun, Feng,
  Liu, Shi, Ouyang, et~al.]{chen2019hybrid}
Kai Chen, Jiangmiao Pang, Jiaqi Wang, Yu~Xiong, Xiaoxiao Li, Shuyang Sun,
  Wansen Feng, Ziwei Liu, Jianping Shi, Wanli Ouyang, et~al.
\newblock Hybrid task cascade for instance segmentation.
\newblock In \emph{CVPR}, pp.\  4974--4983, 2019{\natexlab{a}}.

\bibitem[Chen et~al.(2019{\natexlab{b}})Chen, Wang, Pang, Cao, Xiong, Li, Sun,
  Feng, Liu, Xu, Zhang, Cheng, Zhu, Cheng, Zhao, Li, Lu, Zhu, Wu, Dai, Wang,
  Shi, Ouyang, Loy, and Lin]{mmdetection}
Kai Chen, Jiaqi Wang, Jiangmiao Pang, Yuhang Cao, Yu~Xiong, Xiaoxiao Li,
  Shuyang Sun, Wansen Feng, Ziwei Liu, Jiarui Xu, Zheng Zhang, Dazhi Cheng,
  Chenchen Zhu, Tianheng Cheng, Qijie Zhao, Buyu Li, Xin Lu, Rui Zhu, Yue Wu,
  Jifeng Dai, Jingdong Wang, Jianping Shi, Wanli Ouyang, Chen~Change Loy, and
  Dahua Lin.
\newblock {MMDetection}: Open mmlab detection toolbox and benchmark.
\newblock \emph{arXiv preprint arXiv:1906.07155}, 2019{\natexlab{b}}.

\bibitem[Chen et~al.(2022)Chen, Ge, Tong, Wang, Song, Wang, and
  Luo]{chen2022adaptformer}
Shoufa Chen, Chongjian Ge, Zhan Tong, Jiangliu Wang, Yibing Song, Jue Wang, and
  Ping Luo.
\newblock Adaptformer: Adapting vision transformers for scalable visual
  recognition.
\newblock \emph{arXiv preprint arXiv:2205.13535}, 2022.

\bibitem[Chen et~al.(2021)Chen, Du, Yang, Beyer, Zhai, Lin, Chen, Li, Song,
  Wang, et~al.]{chen2021simple}
Wuyang Chen, Xianzhi Du, Fan Yang, Lucas Beyer, Xiaohua Zhai, Tsung-Yi Lin,
  Huizhong Chen, Jing Li, Xiaodan Song, Zhangyang Wang, et~al.
\newblock A simple single-scale vision transformer for object localization and
  instance segmentation.
\newblock \emph{arXiv preprint arXiv:2112.09747}, 2021.

\bibitem[Cheng et~al.(2021)Cheng, Misra, Schwing, Kirillov, and
  Girdhar]{cheng2021masked}
Bowen Cheng, Ishan Misra, Alexander~G Schwing, Alexander Kirillov, and Rohit
  Girdhar.
\newblock Masked-attention mask transformer for universal image segmentation.
\newblock \emph{arXiv preprint arXiv:2112.01527}, 2021.

\bibitem[Chu et~al.(2021{\natexlab{a}})Chu, Tian, Wang, Zhang, Ren, Wei, Xia,
  and Shen]{chu2021twins}
Xiangxiang Chu, Zhi Tian, Yuqing Wang, Bo~Zhang, Haibing Ren, Xiaolin Wei,
  Huaxia Xia, and Chunhua Shen.
\newblock Twins: Revisiting the design of spatial attention in vision
  transformers.
\newblock \emph{NeurIPS}, 34, 2021{\natexlab{a}}.

\bibitem[Chu et~al.(2021{\natexlab{b}})Chu, Tian, Zhang, Wang, Wei, Xia, and
  Shen]{chu2021conditional}
Xiangxiang Chu, Zhi Tian, Bo~Zhang, Xinlong Wang, Xiaolin Wei, Huaxia Xia, and
  Chunhua Shen.
\newblock Conditional positional encodings for vision transformers.
\newblock \emph{arXiv preprint arXiv:2102.10882}, 2021{\natexlab{b}}.

\bibitem[Contributors(2020)]{mmseg2020}
MMSegmentation Contributors.
\newblock {MMSegmentation}: Openmmlab semantic segmentation toolbox and
  benchmark.
\newblock \url{https://github.com/open-mmlab/mmsegmentation}, 2020.

\bibitem[Deng et~al.(2009)Deng, Dong, Socher, Li, Li, and
  Fei-Fei]{deng2009imagenet}
Jia Deng, Wei Dong, Richard Socher, Li-Jia Li, Kai Li, and Li~Fei-Fei.
\newblock Imagenet: A large-scale hierarchical image database.
\newblock In \emph{CVPR}, pp.\  248--255, 2009.

\bibitem[Dong et~al.(2021)Dong, Bao, Chen, Zhang, Yu, Yuan, Chen, and
  Guo]{dong2021cswin}
Xiaoyi Dong, Jianmin Bao, Dongdong Chen, Weiming Zhang, Nenghai Yu, Lu~Yuan,
  Dong Chen, and Baining Guo.
\newblock Cswin transformer: A general vision transformer backbone with
  cross-shaped windows.
\newblock \emph{arXiv preprint arXiv:2107.00652}, 2021.

\bibitem[Dosovitskiy et~al.(2020)Dosovitskiy, Beyer, Kolesnikov, Weissenborn,
  Zhai, Unterthiner, Dehghani, Minderer, Heigold, Gelly,
  et~al.]{dosovitskiy2020image}
Alexey Dosovitskiy, Lucas Beyer, Alexander Kolesnikov, Dirk Weissenborn,
  Xiaohua Zhai, Thomas Unterthiner, Mostafa Dehghani, Matthias Minderer, Georg
  Heigold, Sylvain Gelly, et~al.
\newblock An image is worth 16x16 words: Transformers for image recognition at
  scale.
\newblock In \emph{ICLR}, 2020.

\bibitem[Fang et~al.(2019)Fang, Sun, Wang, Gou, Li, and Lu]{fang2019instaboost}
Hao-Shu Fang, Jianhua Sun, Runzhong Wang, Minghao Gou, Yong-Lu Li, and Cewu Lu.
\newblock Instaboost: Boosting instance segmentation via probability map guided
  copy-pasting.
\newblock In \emph{ICCV}, pp.\  682--691, 2019.

\bibitem[Fang et~al.(2022)Fang, Yang, Wang, Ge, Shan, and
  Wang]{fang2022unleashing}
Yuxin Fang, Shusheng Yang, Shijie Wang, Yixiao Ge, Ying Shan, and Xinggang
  Wang.
\newblock Unleashing vanilla vision transformer with masked image modeling for
  object detection.
\newblock \emph{arXiv preprint arXiv:2204.02964}, 2022.

\bibitem[Gao et~al.(2021)Gao, Geng, Zhang, Ma, Fang, Zhang, Li, and
  Qiao]{gao2021clip}
Peng Gao, Shijie Geng, Renrui Zhang, Teli Ma, Rongyao Fang, Yongfeng Zhang,
  Hongsheng Li, and Yu~Qiao.
\newblock Clip-adapter: Better vision-language models with feature adapters.
\newblock \emph{arXiv preprint arXiv:2110.04544}, 2021.

\bibitem[Ghiasi et~al.(2021)Ghiasi, Cui, Srinivas, Qian, Lin, Cubuk, Le, and
  Zoph]{ghiasi2021simple}
Golnaz Ghiasi, Yin Cui, Aravind Srinivas, Rui Qian, Tsung-Yi Lin, Ekin~D Cubuk,
  Quoc~V Le, and Barret Zoph.
\newblock Simple copy-paste is a strong data augmentation method for instance
  segmentation.
\newblock In \emph{CVPR}, pp.\  2918--2928, 2021.

\bibitem[Han et~al.(2021)Han, Xiao, Wu, Guo, Xu, and Wang]{han2021transformer}
Kai Han, An~Xiao, Enhua Wu, Jianyuan Guo, Chunjing Xu, and Yunhe Wang.
\newblock Transformer in transformer.
\newblock \emph{NeurIPS}, 34, 2021.

\bibitem[He et~al.(2016)He, Zhang, Ren, and Sun]{he2016deep}
Kaiming He, Xiangyu Zhang, Shaoqing Ren, and Jian Sun.
\newblock Deep residual learning for image recognition.
\newblock In \emph{CVPR}, pp.\  770--778, 2016.

\bibitem[He et~al.(2017)He, Gkioxari, Doll{\'a}r, and Girshick]{he2017mask}
Kaiming He, Georgia Gkioxari, Piotr Doll{\'a}r, and Ross Girshick.
\newblock Mask r-cnn.
\newblock In \emph{ICCV}, pp.\  2961--2969, 2017.

\bibitem[He et~al.(2021)He, Chen, Xie, Li, Doll{\'a}r, and
  Girshick]{he2021masked}
Kaiming He, Xinlei Chen, Saining Xie, Yanghao Li, Piotr Doll{\'a}r, and Ross
  Girshick.
\newblock Masked autoencoders are scalable vision learners.
\newblock \emph{arXiv preprint arXiv:2111.06377}, 2021.

\bibitem[Houlsby et~al.(2019)Houlsby, Giurgiu, Jastrzebski, Morrone,
  De~Laroussilhe, Gesmundo, Attariyan, and Gelly]{houlsby2019parameter}
Neil Houlsby, Andrei Giurgiu, Stanislaw Jastrzebski, Bruna Morrone, Quentin
  De~Laroussilhe, Andrea Gesmundo, Mona Attariyan, and Sylvain Gelly.
\newblock Parameter-efficient transfer learning for nlp.
\newblock In \emph{ICML}, pp.\  2790--2799, 2019.

\bibitem[Huang et~al.(2021{\natexlab{a}})Huang, Lu, Cheng, and
  He]{huang2021fapn}
Shihua Huang, Zhichao Lu, Ran Cheng, and Cheng He.
\newblock Fapn: Feature-aligned pyramid network for dense image prediction.
\newblock In \emph{ICCV}, pp.\  864--873, 2021{\natexlab{a}}.

\bibitem[Huang et~al.(2021{\natexlab{b}})Huang, Ben, Luo, Cheng, Yu, and
  Fu]{huang2021shuffle}
Zilong Huang, Youcheng Ben, Guozhong Luo, Pei Cheng, Gang Yu, and Bin Fu.
\newblock Shuffle transformer: Rethinking spatial shuffle for vision
  transformer.
\newblock \emph{arXiv preprint arXiv:2106.03650}, 2021{\natexlab{b}}.

\bibitem[Jain et~al.(2021)Jain, Singh, Orlov, Huang, Li, Walton, and
  Shi]{jain2021semask}
Jitesh Jain, Anukriti Singh, Nikita Orlov, Zilong Huang, Jiachen Li, Steven
  Walton, and Humphrey Shi.
\newblock Semask: Semantically masked transformers for semantic segmentation.
\newblock \emph{arXiv preprint arXiv:2112.12782}, 2021.

\bibitem[Jia et~al.(2022)Jia, Tang, Chen, Cardie, Belongie, Hariharan, and
  Lim]{jia2022visual}
Menglin Jia, Luming Tang, Bor-Chun Chen, Claire Cardie, Serge Belongie, Bharath
  Hariharan, and Ser-Nam Lim.
\newblock Visual prompt tuning.
\newblock \emph{arXiv preprint arXiv:2203.12119}, 2022.

\bibitem[Jie \& Deng(2022)Jie and Deng]{jie2022convolutional}
Shibo Jie and Zhi-Hong Deng.
\newblock Convolutional bypasses are better vision transformer adapters.
\newblock \emph{arXiv preprint arXiv:2207.07039}, 2022.

\bibitem[Kirillov et~al.(2019)Kirillov, Girshick, He, and
  Doll{\'a}r]{kirillov2019panoptic}
Alexander Kirillov, Ross Girshick, Kaiming He, and Piotr Doll{\'a}r.
\newblock Panoptic feature pyramid networks.
\newblock In \emph{CVPR}, pp.\  6399--6408, 2019.

\bibitem[Li et~al.(2022{\natexlab{a}})Li, Zhang, Liu, Zhang, Ni, Shum,
  et~al.]{li2022maskdino}
Feng Li, Hao Zhang, Shilong Liu, Lei Zhang, Lionel~M Ni, Heung-Yeung Shum,
  et~al.
\newblock Mask dino: Towards a unified transformer-based framework for object
  detection and segmentation.
\newblock \emph{arXiv preprint arXiv:2206.02777}, 2022{\natexlab{a}}.

\bibitem[Li et~al.(2020)Li, Wang, Wu, Chen, Hu, Li, Tang, and
  Yang]{li2020generalized}
Xiang Li, Wenhai Wang, Lijun Wu, Shuo Chen, Xiaolin Hu, Jun Li, Jinhui Tang,
  and Jian Yang.
\newblock Generalized focal loss: Learning qualified and distributed bounding
  boxes for dense object detection.
\newblock \emph{NeurIPS}, 33:\penalty0 21002--21012, 2020.

\bibitem[Li et~al.(2021{\natexlab{a}})Li, Wu, Fan, Mangalam, Xiong, Malik, and
  Feichtenhofer]{li2021improved}
Yanghao Li, Chao-Yuan Wu, Haoqi Fan, Karttikeya Mangalam, Bo~Xiong, Jitendra
  Malik, and Christoph Feichtenhofer.
\newblock Improved multiscale vision transformers for classification and
  detection.
\newblock \emph{arXiv preprint arXiv:2112.01526}, 2021{\natexlab{a}}.

\bibitem[Li et~al.(2021{\natexlab{b}})Li, Xie, Chen, Dollar, He, and
  Girshick]{li2021benchmarking}
Yanghao Li, Saining Xie, Xinlei Chen, Piotr Dollar, Kaiming He, and Ross
  Girshick.
\newblock Benchmarking detection transfer learning with vision transformers.
\newblock \emph{arXiv preprint arXiv:2111.11429}, 2021{\natexlab{b}}.

\bibitem[Li et~al.(2022{\natexlab{b}})Li, Mao, Girshick, and
  He]{li2022exploring}
Yanghao Li, Hanzi Mao, Ross Girshick, and Kaiming He.
\newblock Exploring plain vision transformer backbones for object detection.
\newblock \emph{arXiv preprint arXiv:2203.16527}, 2022{\natexlab{b}}.

\bibitem[Li et~al.(2021{\natexlab{c}})Li, Zhang, Cao, Timofte, and
  Van~Gool]{li2021localvit}
Yawei Li, Kai Zhang, Jiezhang Cao, Radu Timofte, and Luc Van~Gool.
\newblock Localvit: Bringing locality to vision transformers.
\newblock \emph{arXiv preprint arXiv:2104.05707}, 2021{\natexlab{c}}.

\bibitem[Lin et~al.(2014)Lin, Maire, Belongie, Hays, Perona, Ramanan,
  Doll{\'a}r, and Zitnick]{lin2014microsoft}
Tsung-Yi Lin, Michael Maire, Serge Belongie, James Hays, Pietro Perona, Deva
  Ramanan, Piotr Doll{\'a}r, and C~Lawrence Zitnick.
\newblock Microsoft coco: Common objects in context.
\newblock In \emph{ECCV}, pp.\  740--755, 2014.

\bibitem[Liu et~al.(2021{\natexlab{a}})Liu, Hu, Lin, Yao, Xie, Wei, Ning, Cao,
  Zhang, Dong, et~al.]{liu2021swinv2}
Ze~Liu, Han Hu, Yutong Lin, Zhuliang Yao, Zhenda Xie, Yixuan Wei, Jia Ning, Yue
  Cao, Zheng Zhang, Li~Dong, et~al.
\newblock Swin transformer v2: Scaling up capacity and resolution.
\newblock \emph{arXiv preprint arXiv:2111.09883}, 2021{\natexlab{a}}.

\bibitem[Liu et~al.(2021{\natexlab{b}})Liu, Lin, Cao, Hu, Wei, Zhang, Lin, and
  Guo]{liu2021swin}
Ze~Liu, Yutong Lin, Yue Cao, Han Hu, Yixuan Wei, Zheng Zhang, Stephen Lin, and
  Baining Guo.
\newblock Swin transformer: Hierarchical vision transformer using shifted
  windows.
\newblock In \emph{ICCV}, pp.\  10012--10022, 2021{\natexlab{b}}.

\bibitem[Liu et~al.(2021{\natexlab{c}})Liu, Ning, Cao, Wei, Zhang, Lin, and
  Hu]{liu2021video}
Ze~Liu, Jia Ning, Yue Cao, Yixuan Wei, Zheng Zhang, Stephen Lin, and Han Hu.
\newblock Video swin transformer.
\newblock \emph{arXiv preprint arXiv:2106.13230}, 2021{\natexlab{c}}.

\bibitem[Liu et~al.(2022)Liu, Mao, Wu, Feichtenhofer, Darrell, and
  Xie]{liu2022convnet}
Zhuang Liu, Hanzi Mao, Chao-Yuan Wu, Christoph Feichtenhofer, Trevor Darrell,
  and Saining Xie.
\newblock A convnet for the 2020s.
\newblock \emph{arXiv preprint arXiv:2201.03545}, 2022.

\bibitem[Loshchilov \& Hutter(2017)Loshchilov and
  Hutter]{loshchilov2017decoupled}
Ilya Loshchilov and Frank Hutter.
\newblock Decoupled weight decay regularization.
\newblock \emph{arXiv preprint arXiv:1711.05101}, 2017.

\bibitem[Park \& Kim(2022)Park and Kim]{park2022vision}
Namuk Park and Songkuk Kim.
\newblock How do vision transformers work?
\newblock \emph{arXiv preprint arXiv:2202.06709}, 2022.

\bibitem[Peng et~al.(2021)Peng, Huang, Gu, Xie, Wang, Jiao, and
  Ye]{peng2021conformer}
Zhiliang Peng, Wei Huang, Shanzhi Gu, Lingxi Xie, Yaowei Wang, Jianbin Jiao,
  and Qixiang Ye.
\newblock Conformer: Local features coupling global representations for visual
  recognition.
\newblock In \emph{ICCV}, pp.\  367--376, 2021.

\bibitem[Peng et~al.(2022)Peng, Dong, Bao, Ye, and Wei]{peng2022beitv2}
Zhiliang Peng, Li~Dong, Hangbo Bao, Qixiang Ye, and Furu Wei.
\newblock Beit v2: Masked image modeling with vector-quantized visual
  tokenizers.
\newblock \emph{arXiv preprint arXiv:2208.06366}, 2022.

\bibitem[Radford et~al.(2021)Radford, Kim, Hallacy, Ramesh, Goh, Agarwal,
  Sastry, Askell, Mishkin, Clark, et~al.]{radford2021learning}
Alec Radford, Jong~Wook Kim, Chris Hallacy, Aditya Ramesh, Gabriel Goh,
  Sandhini Agarwal, Girish Sastry, Amanda Askell, Pamela Mishkin, Jack Clark,
  et~al.
\newblock Learning transferable visual models from natural language
  supervision.
\newblock In \emph{International Conference on Machine Learning}, pp.\
  8748--8763. PMLR, 2021.

\bibitem[Ranftl et~al.(2021)Ranftl, Bochkovskiy, and Koltun]{ranftl2021vision}
Ren{\'e} Ranftl, Alexey Bochkovskiy, and Vladlen Koltun.
\newblock Vision transformers for dense prediction.
\newblock In \emph{ICCV}, pp.\  12179--12188, 2021.

\bibitem[Rao et~al.(2022)Rao, Zhao, Tang, Zhou, Lim, and Lu]{rao2022hornet}
Yongming Rao, Wenliang Zhao, Yansong Tang, Jie Zhou, Ser-Nam Lim, and Jiwen Lu.
\newblock Hornet: Efficient high-order spatial interactions with recursive
  gated convolutions.
\newblock \emph{arXiv preprint arXiv:2207.14284}, 2022.

\bibitem[Rebuffi et~al.(2017)Rebuffi, Bilen, and Vedaldi]{rebuffi2017learning}
Sylvestre-Alvise Rebuffi, Hakan Bilen, and Andrea Vedaldi.
\newblock Learning multiple visual domains with residual adapters.
\newblock \emph{NeurIPS}, 30, 2017.

\bibitem[Rebuffi et~al.(2018)Rebuffi, Bilen, and Vedaldi]{rebuffi2018efficient}
Sylvestre-Alvise Rebuffi, Hakan Bilen, and Andrea Vedaldi.
\newblock Efficient parametrization of multi-domain deep neural networks.
\newblock In \emph{CVPR}, pp.\  8119--8127, 2018.

\bibitem[Rosenfeld \& Tsotsos(2018)Rosenfeld and
  Tsotsos]{rosenfeld2018incremental}
Amir Rosenfeld and John~K Tsotsos.
\newblock Incremental learning through deep adaptation.
\newblock \emph{TPAMI}, 42\penalty0 (3):\penalty0 651--663, 2018.

\bibitem[Shao et~al.(2019)Shao, Li, Zhang, Peng, Yu, Zhang, Li, and
  Sun]{shao2019objects365}
Shuai Shao, Zeming Li, Tianyuan Zhang, Chao Peng, Gang Yu, Xiangyu Zhang, Jing
  Li, and Jian Sun.
\newblock Objects365: A large-scale, high-quality dataset for object detection.
\newblock In \emph{ICCV}, pp.\  8430--8439, 2019.

\bibitem[Si et~al.(2022)Si, Yu, Zhou, Zhou, Wang, and Yan]{si2022inception}
Chenyang Si, Weihao Yu, Pan Zhou, Yichen Zhou, Xinchao Wang, and Shuicheng Yan.
\newblock Inception transformer.
\newblock \emph{arXiv preprint arXiv:2205.12956}, 2022.

\bibitem[Steiner et~al.(2021)Steiner, Kolesnikov, Zhai, Wightman, Uszkoreit,
  and Beyer]{steiner2021train}
Andreas Steiner, Alexander Kolesnikov, Xiaohua Zhai, Ross Wightman, Jakob
  Uszkoreit, and Lucas Beyer.
\newblock How to train your vit? data, augmentation, and regularization in
  vision transformers.
\newblock \emph{arXiv preprint arXiv:2106.10270}, 2021.

\bibitem[Stickland \& Murray(2019)Stickland and Murray]{stickland2019bert}
Asa~Cooper Stickland and Iain Murray.
\newblock Bert and pals: Projected attention layers for efficient adaptation in
  multi-task learning.
\newblock In \emph{ICML}, pp.\  5986--5995, 2019.

\bibitem[Strudel et~al.(2021)Strudel, Garcia, Laptev, and
  Schmid]{strudel2021segmenter}
Robin Strudel, Ricardo Garcia, Ivan Laptev, and Cordelia Schmid.
\newblock Segmenter: Transformer for semantic segmentation.
\newblock In \emph{ICCV}, pp.\  7262--7272, 2021.

\bibitem[Sung et~al.(2021)Sung, Cho, and Bansal]{sung2021vl}
Yi-Lin Sung, Jaemin Cho, and Mohit Bansal.
\newblock Vl-adapter: Parameter-efficient transfer learning for
  vision-and-language tasks.
\newblock \emph{arXiv preprint arXiv:2112.06825}, 2021.

\bibitem[Touvron et~al.(2021)Touvron, Cord, Douze, Massa, Sablayrolles, and
  J{\'e}gou]{touvron2021training}
Hugo Touvron, Matthieu Cord, Matthijs Douze, Francisco Massa, Alexandre
  Sablayrolles, and Herv{\'e} J{\'e}gou.
\newblock Training data-efficient image transformers \& distillation through
  attention.
\newblock In \emph{ICML}, pp.\  10347--10357, 2021.

\bibitem[Vaswani et~al.(2017)Vaswani, Shazeer, Parmar, Uszkoreit, Jones, Gomez,
  Kaiser, and Polosukhin]{vaswani2017attention}
Ashish Vaswani, Noam Shazeer, Niki Parmar, Jakob Uszkoreit, Llion Jones,
  Aidan~N Gomez, {\L}ukasz Kaiser, and Illia Polosukhin.
\newblock Attention is all you need.
\newblock \emph{NeurIPS}, 30, 2017.

\bibitem[Wang et~al.(2021)Wang, Xie, Li, Fan, Song, Liang, Lu, Luo, and
  Shao]{wang2021pyramid}
Wenhai Wang, Enze Xie, Xiang Li, Deng-Ping Fan, Kaitao Song, Ding Liang, Tong
  Lu, Ping Luo, and Ling Shao.
\newblock Pyramid vision transformer: A versatile backbone for dense prediction
  without convolutions.
\newblock In \emph{ICCV}, pp.\  568--578, 2021.

\bibitem[Wang et~al.(2022{\natexlab{a}})Wang, Xie, Li, Fan, Song, Liang, Lu,
  Luo, and Shao]{wang2021pvtv2}
Wenhai Wang, Enze Xie, Xiang Li, Deng-Ping Fan, Kaitao Song, Ding Liang, Tong
  Lu, Ping Luo, and Ling Shao.
\newblock Pvtv2: Improved baselines with pyramid vision transformer.
\newblock \emph{CVMJ}, pp.\  1--10, 2022{\natexlab{a}}.

\bibitem[Wang et~al.(2022{\natexlab{b}})Wang, Bao, Dong, Bjorck, Peng, Liu,
  Aggarwal, Mohammed, Singhal, Som, et~al.]{wang2022beit3}
Wenhui Wang, Hangbo Bao, Li~Dong, Johan Bjorck, Zhiliang Peng, Qiang Liu, Kriti
  Aggarwal, Owais~Khan Mohammed, Saksham Singhal, Subhojit Som, et~al.
\newblock Image as a foreign language: Beit pretraining for all vision and
  vision-language tasks.
\newblock \emph{arXiv preprint arXiv:2208.10442}, 2022{\natexlab{b}}.

\bibitem[Wei et~al.(2022)Wei, Hu, Xie, Zhang, Cao, Bao, Chen, and
  Guo]{wei2022kdswin}
Yixuan Wei, Han Hu, Zhenda Xie, Zheng Zhang, Yue Cao, Jianmin Bao, Dong Chen,
  and Baining Guo.
\newblock Contrastive learning rivals masked image modeling in fine-tuning via
  feature distillation.
\newblock \emph{arXiv preprint arXiv:2205.14141}, 2022.

\bibitem[Wu et~al.(2021)Wu, Xiao, Codella, Liu, Dai, Yuan, and
  Zhang]{wu2021cvt}
Haiping Wu, Bin Xiao, Noel Codella, Mengchen Liu, Xiyang Dai, Lu~Yuan, and Lei
  Zhang.
\newblock Cvt: Introducing convolutions to vision transformers.
\newblock In \emph{ICCV}, pp.\  22--31, 2021.

\bibitem[Wu et~al.(2022{\natexlab{a}})Wu, Wu, Tan, and Guo]{wu2022pale}
Sitong Wu, Tianyi Wu, Haoru Tan, and Guodong Guo.
\newblock Pale transformer: A general vision transformer backbone with
  pale-shaped attention.
\newblock In \emph{AAAI}, volume~36, pp.\  2731--2739, 2022{\natexlab{a}}.

\bibitem[Wu et~al.(2022{\natexlab{b}})Wu, Liu, Zhan, and Cheng]{wu2022p2t}
Yu-Huan Wu, Yun Liu, Xin Zhan, and Ming-Ming Cheng.
\newblock P2t: Pyramid pooling transformer for scene understanding.
\newblock \emph{TPAMI}, 2022{\natexlab{b}}.

\bibitem[Xiao et~al.(2018)Xiao, Liu, Zhou, Jiang, and Sun]{xiao2018unified}
Tete Xiao, Yingcheng Liu, Bolei Zhou, Yuning Jiang, and Jian Sun.
\newblock Unified perceptual parsing for scene understanding.
\newblock In \emph{ECCV}, pp.\  418--434, 2018.

\bibitem[Xie et~al.(2021)Xie, Wang, Yu, Anandkumar, Alvarez, and
  Luo]{xie2021segformer}
Enze Xie, Wenhai Wang, Zhiding Yu, Anima Anandkumar, Jose~M Alvarez, and Ping
  Luo.
\newblock Segformer: Simple and efficient design for semantic segmentation with
  transformers.
\newblock \emph{NeurIPS}, 34, 2021.

\bibitem[Yang et~al.(2021)Yang, Li, Zhang, Dai, Xiao, Yuan, and
  Gao]{yang2021focal}
Jianwei Yang, Chunyuan Li, Pengchuan Zhang, Xiyang Dai, Bin Xiao, Lu~Yuan, and
  Jianfeng Gao.
\newblock Focal self-attention for local-global interactions in vision
  transformers.
\newblock \emph{arXiv preprint arXiv:2107.00641}, 2021.

\bibitem[Zhang et~al.(2021)Zhang, Fang, Gao, Zhang, Li, Dai, Qiao, and
  Li]{zhang2021tip}
Renrui Zhang, Rongyao Fang, Peng Gao, Wei Zhang, Kunchang Li, Jifeng Dai,
  Yu~Qiao, and Hongsheng Li.
\newblock Tip-adapter: Training-free clip-adapter for better vision-language
  modeling.
\newblock \emph{arXiv preprint arXiv:2111.03930}, 2021.

\bibitem[Zhang et~al.(2020)Zhang, Chi, Yao, Lei, and Li]{zhang2020bridging}
Shifeng Zhang, Cheng Chi, Yongqiang Yao, Zhen Lei, and Stan~Z Li.
\newblock Bridging the gap between anchor-based and anchor-free detection via
  adaptive training sample selection.
\newblock In \emph{CVPR}, pp.\  9759--9768, 2020.

\bibitem[Zhang et~al.(2022)Zhang, Zhou, and Liu]{zhang2022neural}
Yuanhan Zhang, Kaiyang Zhou, and Ziwei Liu.
\newblock Neural prompt search.
\newblock \emph{arXiv preprint arXiv:2206.04673}, 2022.

\bibitem[Zheng et~al.(2021)Zheng, Lu, Zhao, Zhu, Luo, Wang, Fu, Feng, Xiang,
  Torr, et~al.]{zheng2021rethinking}
Sixiao Zheng, Jiachen Lu, Hengshuang Zhao, Xiatian Zhu, Zekun Luo, Yabiao Wang,
  Yanwei Fu, Jianfeng Feng, Tao Xiang, Philip~HS Torr, et~al.
\newblock Rethinking semantic segmentation from a sequence-to-sequence
  perspective with transformers.
\newblock In \emph{CVPR}, pp.\  6881--6890, 2021.

\bibitem[Zhou et~al.(2017)Zhou, Zhao, Puig, Fidler, Barriuso, and
  Torralba]{zhou2017scene}
Bolei Zhou, Hang Zhao, Xavier Puig, Sanja Fidler, Adela Barriuso, and Antonio
  Torralba.
\newblock Scene parsing through ade20k dataset.
\newblock In \emph{CVPR}, pp.\  633--641, 2017.

\bibitem[Zhu et~al.(2022)Zhu, Zhu, Wang, Wang, Li, Wang, and Dai]{zhu2022uni}
Jinguo Zhu, Xizhou Zhu, Wenhai Wang, Xiaohua Wang, Hongsheng Li, Xiaogang Wang,
  and Jifeng Dai.
\newblock Uni-perceiver-moe: Learning sparse generalist models with conditional
  moes.
\newblock \emph{arXiv preprint arXiv:2206.04674}, 2022.

\bibitem[Zhu et~al.(2020)Zhu, Su, Lu, Li, Wang, and Dai]{zhu2020deformable}
Xizhou Zhu, Weijie Su, Lewei Lu, Bin Li, Xiaogang Wang, and Jifeng Dai.
\newblock Deformable detr: Deformable transformers for end-to-end object
  detection.
\newblock In \emph{ICLR}, 2020.

\bibitem[Zhu et~al.(2021)Zhu, Zhu, Li, Wu, Wang, Li, Wang, and Dai]{zhu2021uni}
Xizhou Zhu, Jinguo Zhu, Hao Li, Xiaoshi Wu, Xiaogang Wang, Hongsheng Li,
  Xiaohua Wang, and Jifeng Dai.
\newblock Uni-perceiver: Pre-training unified architecture for generic
  perception for zero-shot and few-shot tasks.
\newblock \emph{arXiv preprint arXiv:2112.01522}, 2021.

\end{thebibliography}
\bibliographystyle{iclr2023_conference}

\newpage

\appendix
\section{Comparison with Previous State-of-the-Arts}

In recent years, the state-of-the-art models on dense prediction benchmarks are primarily vision-specific transformers, such as Swin~\citep{liu2021swin}, Focal~\citep{yang2021focal}, MViTv2~\citep{li2021improved}, and SwinV2~\citep{liu2021swinv2}, while the plain ViT is rarely found.
Nevertheless, we argue that the plain ViT still has the potential to reach the leading performance by leveraging our ViT-Adapter.
To verify this, we conduct extensive additional experiments as follows.

\begin{table}[t]\small
	\centering
	\renewcommand\arraystretch{1.0}
    \setlength\tabcolsep{0.85mm}
    \begin{tabular}{l|c|c|c|cc|cc|cc|cc}
        \toprule
        \multirow{2}{*}{Method} & \multirow{2}{*}{Framework} & \multirow{2}{*}{Epoch} & Backbone & \multicolumn{2}{c|}{val}  &\multicolumn{2}{c|}{val (+MS)}  &\multicolumn{2}{c|}{test-dev}  & \multicolumn{2}{c}{test-dev~(+MS)} \\
         & & & Pre-train & AP$^{\rm b}$ & AP$^{\rm m}$ & AP$^{\rm b}$ & AP$^{\rm m}$& AP$^{\rm b}$ & AP$^{\rm m}$& ~~~AP$^{\rm b}$~ & AP$^{\rm m}$ \\
        \midrule
        Swin-L & HTC++ & 72 & IN-22K, sup & 57.1 & 49.5 & 58.0 & 50.4 & 57.7 & 50.2 & 58.7 & 51.1 \\
        Focal-L & HTC++ & 36 & IN-22K, sup & 57.0 & 49.9 & 58.1 & 50.9 & - & - & 58.4 & 51.3  \\
        MViTv2-L & Cascade & 50 & IN-22K, sup &56.9 & 48.6 & 58.7 & 50.5 & - & - & - & - \\
        MViTv2-H & Cascade & 50 & IN-22K, sup & 57.1 & 48.8 & 58.4 & 50.1 & - & - & - & - \\
        CBV2-Swin-L & HTC  & 36& IN-22K, sup & 59.1 & 51.0 & 59.6 & 51.8 & 59.4 & 51.6 & 60.1 & 52.3  \\
	    \rowcolor{gray!20} 
        ViT-Adapter-L & HTC++ & 36 & IN-22K, sup & 56.6 & 49.0 & 57.7 & 49.9 & 57.4 & 50.0 & 58.4 & 50.7 \\
        \midrule
        Swin-L & HTC++ & 36 & IN-1K, UM-MAE & 57.4 & 49.8 & 58.7 & 50.9 & - & - & - & - \\
        ViTDet-L & Cascade & 100 & IN-1K, MAE & \textbf{59.6} & 51.1 & 60.4 & 52.2 & -& -& -& -\\
        \rowcolor{gray!20} 
        ViT-Adapter-L  & HTC++ & 36& IN-22K, BEiT & 58.4 & 50.8 & 60.2 & 52.2 & 58.9 & 51.3 & 60.4 & 52.5  \\
        \rowcolor{yellow!15} 
        ViT-Adapter-L  & HTC++ & 36 & MM$^\dagger$, BEiTv2 & 58.8& \textbf{51.1} & \textbf{60.5} & \textbf{52.5} & \textbf{59.5} & \textbf{51.8} & \textbf{60.9} & \textbf{53.0}  \\
	    \bottomrule
    \end{tabular}
    \caption{\textbf{Comparisons with the leading results on the COCO val2017 and test-dev sets.} 
    There are three detection frameworks used, including Cascade Mask R-CNN~\citep{cai2019cascade}, HTC~\citep{chen2019hybrid}, and its extension HTC++~\citep{liu2021swin}.
    All of these models are trained \emph{without} extra detection datasets, such as Objects365~\citep{shao2019objects365}.
    ``IN-22K, sup'' is short for ImageNet-22K supervised pre-training.
    ``MS" indicates multi-scale testing.
    ``$^\dagger$": Since the pre-trained CLIP~\citep{radford2021learning} model is used in the training process of BEiTv2~\citep{peng2022beitv2}, we regard it as a multi-modal pre-training method.
    }
    \label{tab:sota_coco}
\end{table}

\subsection{Object Detection and Instance Segmentation}
\label{sec:appendix_detection}

\noindent \textbf{Settings.}
Following prior art~\citep{li2021benchmarking}, we modify the 24-layer ViT-L to use 14$\times$14 window attention except for layers spaced at an interval of 6, to save training time and memory. 
The state-of-the-art detector HTC++~\citep{liu2021swin} is employed for our experiments.
Specifically, we rescale the shorter side of images between 400 and 1400, while the longer side is at most 1600. Instaboost~\citep{fang2019instaboost}, Soft-NMS~\citep{bodla2017soft}, AdamW~\citep{loshchilov2017decoupled} optimizer
(batch size of 16, initial learning rate of 1$\times$10$^{-4}$, and weight decay of 0.05), and 3$\times$ schedule are adopted during training.
We use a layer-wise learning rate decay of 0.9, and a drop path rate of 0.4. 
For a fairer comparison, here we take two initialization strategies, \ie~regular ImageNet-22K pre-training, and more advanced self-supervised or multi-modal pre-training.

\noindent \textbf{Results with ImageNet-22K Pre-training.}
As shown in Table~\ref{tab:sota_coco}, 
with the ImageNet-22K supervised pre-training from AugReg~\citep{steiner2021train},
our ViT-Adapter-L reports 58.4 AP$\rm^{b}$ and 50.7 AP$\rm^{m}$ on the COCO test-dev, which is
comparable to many vision-specific transformers such as Swin-L (58.4 AP$\rm^b$ \emph{vs.}~58.7 AP$\rm^b$) and Focal-L (58.4 AP$\rm^b$ \emph{vs.}~58.4 AP$\rm^b$).
This fair comparison illustrates that, our ViT-Adapter significantly narrows the performance gap between the plain ViT and well-designed vision-specific models.

\noindent \textbf{Results with More Advanced Pre-training.}
Since our paradigm retains the flexibility of the plain ViT, it can \emph{easily derive significant benefits from advanced pre-training techniques}, such as multi-modal pre-training~\citep{zhu2021uni,zhu2022uni} or self-supervised pre-training~\citep{bao2021beit,peng2022beitv2,he2021masked}, or a combination of the both~\citep{wang2022beit3}.
Here, we take the readily available weights from BEiT~\citep{bao2021beit} and BEiTv2~\citep{bao2021beit} as examples. 
Due to BEiT using learnable relative position biases instead of the absolute position embeddings, we replace the remaining global attention (see the settings part) with 56$\times$56 window attention as an approximation.
For these layers, the relative position biases need to be interpolated to adapt to the new window size.

As reported in Table~\ref{tab:sota_coco},
our ViT-Adapter-L (w/ BEiT) creates 60.4 AP$\rm^b$ and 52.5 AP$\rm^m$ on the COCO test-dev, and ViT-Adapter-L (w/ BEiTv2) further sets this record to 60.9 AP$\rm^b$ and 53.0 AP$\rm^m$.
Notably, although it's not a perfectly controlled comparison, our method attains similar performance taking fewer training epochs (36 \emph{vs.}~100) than ViTDet~\citep{li2022exploring}.
We argue that a longer training schedule such as 100 epochs may bring an added bonus, but it is expensive to afford due to limited computing resources.
In summary, from a system-level perspective, our ViT-Adapter can enjoy the dividends of various advanced pre-training techniques and help plain ViT achieve leading performance on the object detection and instance segmentation tasks.

\begin{table}[t]\small
	\centering
	\renewcommand\arraystretch{1.0}
    \setlength\tabcolsep{0.40mm}
    \begin{tabular}{l|c|c|c|c|c|cc|c}
        \toprule
        \multirow{2}{*}{Method} & \multirow{2}{*}{Framework} & Backbone & Extra & Crop & \multirow{2}{*}{Iters} & \multicolumn{2}{c|}{ADE20K val} & \multirow{2}{*}{\#Param}\\
        & & Pre-train & Pre-train& Size & & mIoU & +MS & \\
        \midrule
        Swin-L & Mask2Former & IN-22K, sup & - & 640 & 160k & 56.1 & 57.3 & 215M\\
        Swin-L-FaPN & Mask2Former & IN-22K, sup & -& 640 & 160k & 56.4 & 57.7 & 217M\\
        SeMask-Swin-L & Mask2Former & IN-22K, sup & -& 640 & 160k & 57.0 & 58.2 & -\\
        HorNet-L & Mask2Former & IN-22K, sup & -& 640 & 160k & 57.5 & 57.9 & -\\
        \rowcolor{gray!20} 
        ViT-Adapter-L & Mask2Former & IN-22K, sup & - & 640 & 160k & 56.8 & 57.7 & 438M \\
        \midrule
	    BEiT-L & UperNet & IN-22K, BEiT & - & 640 & 160k & 56.7 & 57.0 & 441M\\
	    \rowcolor{gray!20} 
	    ViT-Adapter-L &
	    UperNet & IN-22K, BEiT & - & 640 & 160k & 58.0 & 58.4 & 451M\\
	    BEiTv2-L & UperNet & IN-22K, BEiTv2 & - & 512 & 160k & 57.5 & 58.0 & 441M\\
	    \rowcolor{gray!20} 
	    ViT-Adapter-L &
	    UperNet & IN-22K, BEiTv2 & - & 512 & 160k & 58.0 & 58.5  & 451M\\
	   \midrule
	   ConvNeXt-XL* & Mask2Former & IN-22K, sup & COCO-Stuff, sup & 896 & 80k & 57.1 & 58.4 & 588M \\
	   Swin-L* & Mask2Former & IN-22K, sup & COCO-Stuff, sup & 896  & 80k & 57.3 & 58.3 & 434M \\
	   SwinV2-G & UperNet & IN-22K, sup & Ext-70M, sup & 896 & 160k & 59.3 & 59.9 & 3.0B \\
	   FD-SwinV2-G & UperNet & IN-22K, sup & Ext-70M, sup & 896 & 160k & - & 61.4 & 3.0B \\
	   Swin-L & Mask DINO & IN-22K, sup & Objects365, sup & - & 160k &59.5 & 60.8 & 223M  \\
	    \rowcolor{gray!20} 
	    ViT-Adapter-L & Mask2Former &IN-22K, BEiT & COCO-Stuff, sup & 896 & 80k & 59.4 & 60.5 & 571M\\
	    \rowcolor{yellow!15} 
	    ViT-Adapter-L & Mask2Former &MM$^\dagger$, BEiTv2 & COCO-Stuff, sup & 896 & 80k & \textbf{61.2} & \textbf{61.5} & 571M\\
	    \rowcolor{yellow!15} 
	    BEiT-3 (w/ ViT-Adapter) & Mask2Former & MM, BEiT-3 & COCO-Stuff, sup & 896 & 80k  & \textbf{62.0} & \textbf{62.8} & 1.3B\\
	    \bottomrule
    \end{tabular}
    \caption{\textbf{Comparison with previous state-of-the-art results on the ADE20K validation set.} 
	The results of BEiT-3 are collected from~\citep{wang2022beit3}.
	``$^*$'': We follow BEiT~\citep{bao2021beit} to use a wider segmentation head for ConvNeXt-XL and Swin-L to match the number of parameters, and apply the same training strategy to them.
    ``IN-22K, sup'': ImageNet-22K supervised pre-training.
    ``MM": Multi-modal pre-training.
    ``MS": Multi-scale testing.
    ``$^\dagger$": Since the pre-trained CLIP~\citep{radford2021learning} model is used in the training process of BEiTv2~\citep{peng2022beitv2}, we regard it as a multi-modal pre-training method.
    }
    \label{tab:sota_ade}
\end{table}

\subsection{Semantic Segmentation}
\label{sec:appendix_segmentation}

\noindent\textbf{Settings.} For semantic segmentation, we employ the AdamW optimizer with an initial learning rate of 2$\times$10$^{-5}$, a batch size of 16, and a weight decay of 0.05. 
Layer-wise learning rate decay of 0.9 and drop path rate of 0.4 are used to train the models. 
Other training settings, such as pre-training techniques, crop size, and the number of iterations, are listed in Table~\ref{tab:sota_ade}.

\noindent \textbf{Results with ImageNet-22K Pre-training.}
As shown in Table~\ref{tab:sota_ade}, 
using Mask2Former~\citep{cheng2021masked} as the segmenter,
our ViT-Adapter-L achieves 56.8 mIoU and 57.7 MS mIoU on the ADE20K val, which is comparable to recent vision-specific models, such as Swin-L~\citep{liu2021swin}, Swin-L-FaPN~\citep{huang2021fapn}, SeMask-Swin-L~\citep{jain2021semask}, and HorNet-L~\citep{rao2022hornet}.

\noindent \textbf{Results with More Advanced Pre-training.}
It can be seen from Table~\ref{tab:sota_ade}, when training with UperNet for 160k iterations, our ViT-Adapter-L (w/ BEiT) yields 58.4 MS mIoU, outperforming BEiT-L by 1.4 points with only 10M additional parameters.
It shows that our adapter can deliver significant benefits even for a powerful self-supervised pre-trained ViT.

Furthermore, we compare the performance of our method with vision-specific models that also use additional datasets.
For example, SwinV2-G~\citep{liu2021swinv2} uses a privately collected ImageNet-22K-ext-70M dataset that contains 70 million images.
Mask DINO~\citep{li2022maskdino} takes the detection pre-training on large-scale Objects365~\citep{shao2019objects365} dataset as the initialization for segmentation.
Due to limited computing resources, we explore a \emph{simple and affordable} transfer learning strategy for semantic segmentation.
Specifically, we use the COCO-Stuff~\citep{caesar2018coco} dataset for 80k iterations of pre-training, and then ADE20K for 80k iterations of fine-tuning.
The total number of iterations is still 160k, and no additional training overhead is added.

Under this setting, our ViT-Adapter-L (w/ BEiT) produces an exciting score of 60.5 MS mIoU. Further, ViT-Adapter-L (w/ BEiTv2) creates a new record of 61.5 MS mIoU, which is slightly better than FD-SwinV2-G~\citep{wei2022kdswin}, while the parameter number is much smaller (571M \emph{vs.}~3.0B). 
It's worth noting that, our ViT-Adapter is also adopted by the recently proposed BEiT-3~\citep{wang2022beit3}, which is a ViT-style foundation model that can be pre-trained with multi-modal data.
As described in their paper, using ViT-Adapter for the transfer learning of semantic segmentation,
BEiT-3 establishes a new state-of-the-art of 62.8 MS mIoU on ADE20K val, which is a convincing verification of the paradigm we present in Figure~\ref{fig:paradigm}.

\section{Additional Ablation and Discussion}

\begin{table}[t]\small
	\centering
	\renewcommand\arraystretch{1.0}
    \setlength\tabcolsep{2.3mm}
    \begin{tabular}{l|ccccc|cccc|c}
        \toprule
        \multirow{2}{*}{Variants} & \multicolumn{5}{c|}{Settings of ViT} & \multicolumn{4}{c|}{Settings of Adapter} & Total\\
         & Layers & Width & FFN & Heads & \#Param & $N$ & FFN & Heads & \#Param& Param\\
	    \midrule
        Tiny (T)   & 12 & 192  & 768  & 3   & 5.5M   & 4 & 48  & 6  & 2.5M & 8.0M \\ 
        Small (S)  & 12 & 384  & 1536 & 6   & 21.7M  & 4 & 96  & 6  & 5.8M & 27.5M \\ 
        Base (B)   & 12 & 768  & 3072 & 12  & 85.8M  & 4 & 192 & 12 & 14.0M & 99.8M \\
        Large (L)  & 24 & 1024 & 4096 & 16  & 303.3M & 4 & 256 & 16 & 23.7M & 327.0M \\
	    \bottomrule
    \end{tabular}
    \caption{\textbf{Configurations of the ViT-Adapter.} 
    We apply our adapters on four different settings of ViT, including ViT-T, ViT-S, ViT-B, and ViT-L, covering a wide range of different model sizes.}
    \label{tab:architecture_configurations}
\end{table}

\noindent \textbf{Architecture Configurations.}
\label{appendix_config}
The more detailed configurations are listed in Table~\ref{tab:architecture_configurations}.


\begin{figure}[t]
    \centering
    \includegraphics[width=1\linewidth]{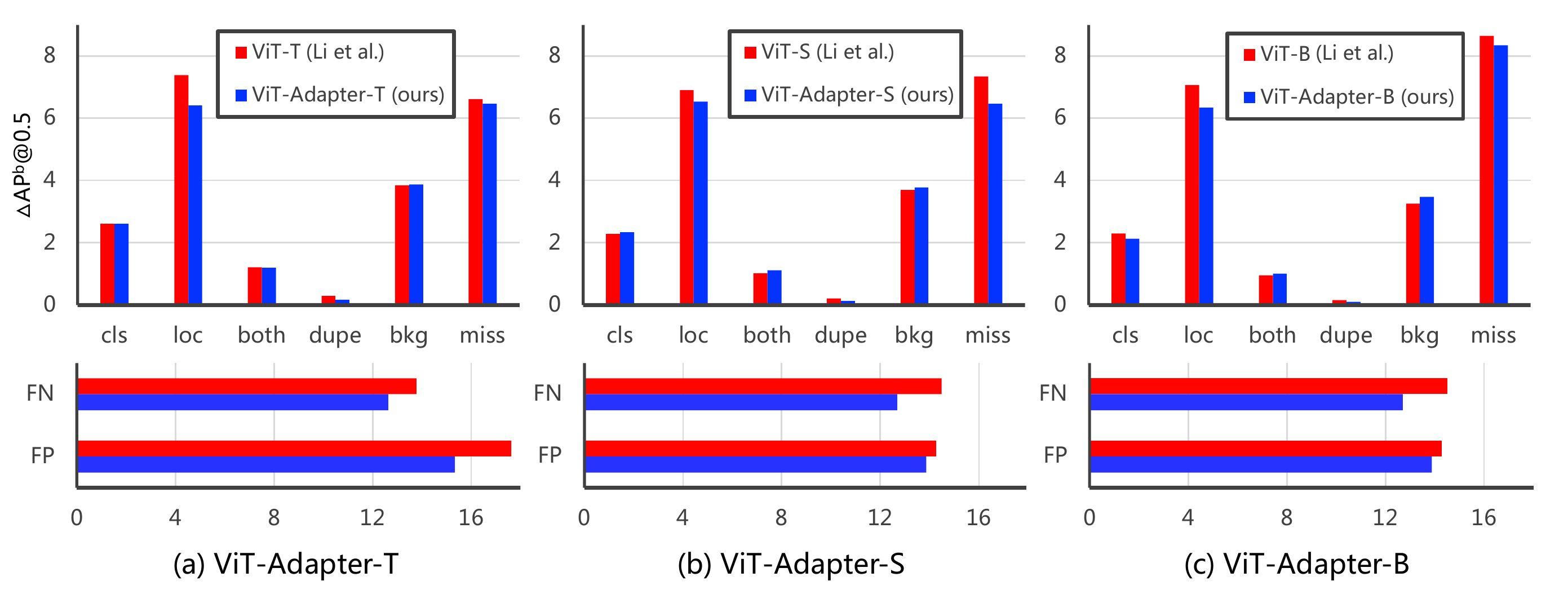}
    \caption{
    \textbf{TIDE error type analysis (the lower the better). }
    We use the models listed in Table~\ref{tab:results_detection_mask} for analysis.
    As defined in~\citep{bolya2020tide}, we plot the AP$^\text{b}$ metric at an IoU threshold of 0.5.
    These bars show the effect of each error type on overall detection performance).
    The error types include: 
    \textbf{cls}: localized correctly but classified incorrectly; 
    \textbf{loc}: classified correctly but localized incorrectly; 
    \textbf{both}: classified incorrectly and localized incorrectly; 
    \textbf{dupe}: detection would be correct if not for a higher scoring detection; 
    \textbf{bkg}: detected background as foreground; 
    \textbf{miss}: all undetected ground-truth not covered by other error types;
    \textbf{FN}: false negatives;
    \textbf{FP}: false positives.
    We observe that our ViT-Adapter makes fewer localization and miss errors than the ViT baseline~\citep{li2021benchmarking}, and occurs fewer false positive and negative errors.
        }
    \label{fig:tide}
\end{figure}

\noindent\textbf{TIDE Error Type Analysis.}
TIDE~\citep{bolya2020tide} is a toolbox for analyzing the sources of error in object detection algorithms.
Following~\citep{li2021benchmarking}, we show the error type analysis in Figure~\ref{fig:tide}.
For fair comparison, the models listed in Table~\ref{tab:results_detection_mask} are adopted for analysis.
These results reveal where our ViT-Adapter improves overall AP$^\text{b}$ relative to the ViT baseline~\citep{li2021benchmarking}.
For instance, we observe that our adapter helps reduce missed and localization errors, and has a substantial effect on fixing false negative and positive errors.

\begin{figure}[t]
    \centering
    \includegraphics[width=1\linewidth]{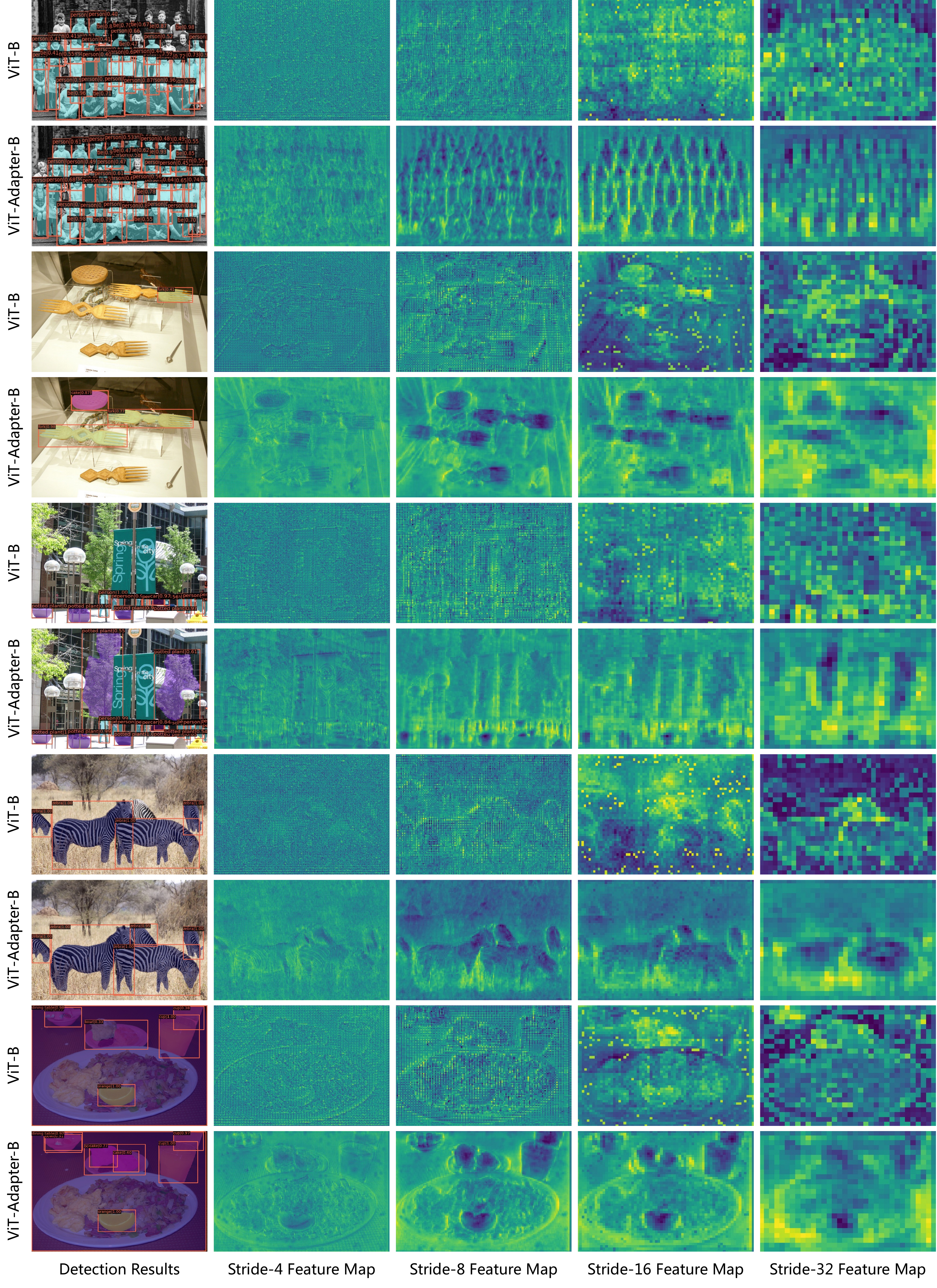}
    \caption{
        \textbf{Visualization of feature maps for object detection and instance segmentation. }
        Compared to the ViT baseline~\citep{li2021benchmarking}, our ViT-Adapter yields more fine-grained multi-scale feature maps, thus improving localization quality and reducing missed detection.
        }
    \label{fig:feature_map_1}
\end{figure}

\begin{figure}[t]
    \centering
    \includegraphics[width=1\linewidth]{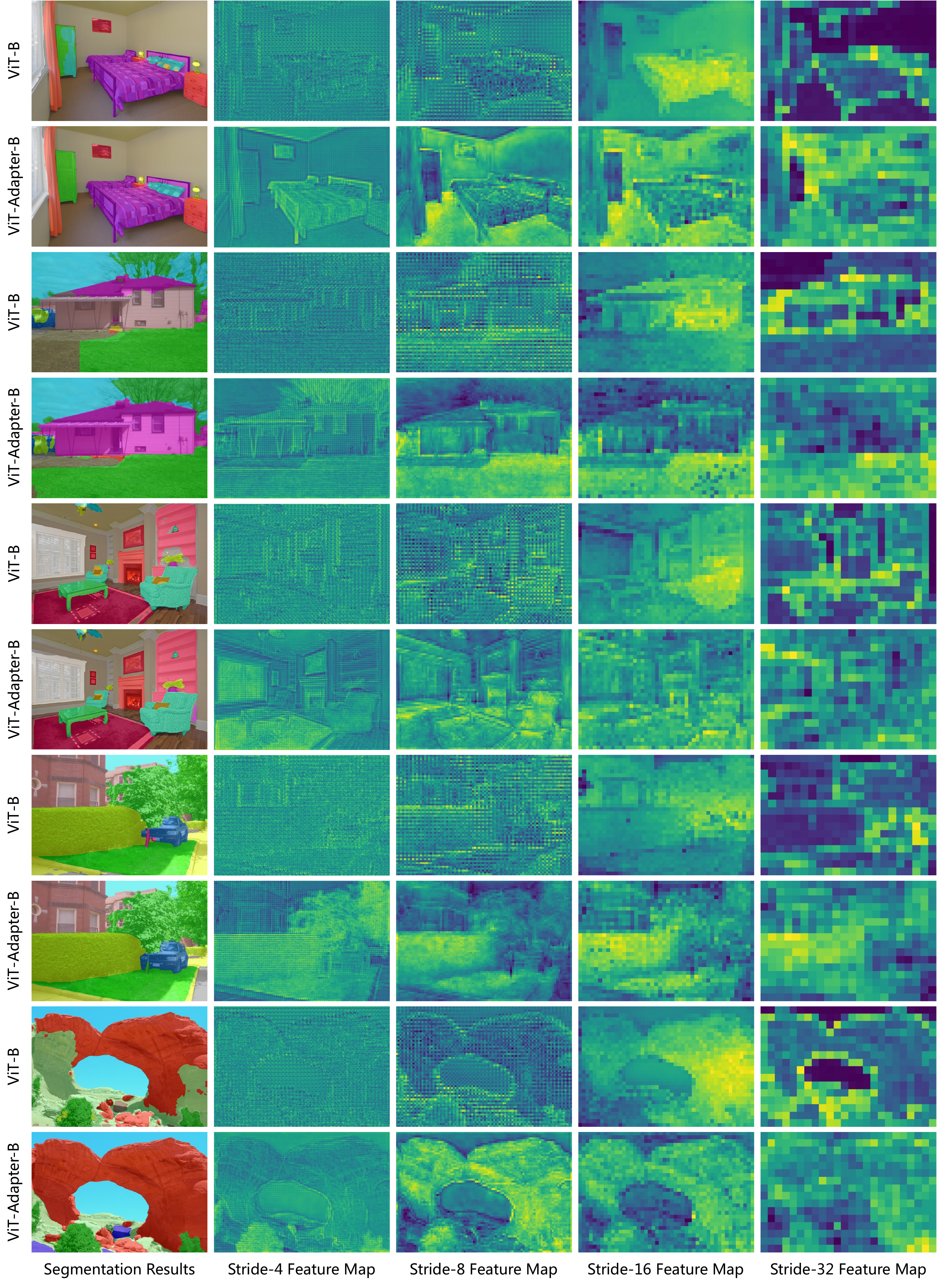}
    \caption{
        \textbf{Visualization of feature maps for semantic segmentation. }
        Compared to the ViT baseline~\citep{li2021benchmarking}, our ViT-Adapter yields more fine-grained multi-scale feature maps with rich local edges and textures, thus improving the performance of semantic segmentation.
        }
    \label{fig:feature_map_2}
\end{figure}

\noindent\textbf{Feature Visualization.}
We plot more visualization of feature maps produced by ViT-B \citep{li2021benchmarking} and our ViT-Adapter-B in Figure~\ref{fig:feature_map_1} and Figure~\ref{fig:feature_map_2}, which are trained based on Mask R-CNN for detection and UperNet for segmentation, respectively.
As can be seen, the features of ViT-B are blurry and coarse, while our features are more refined and have more local edges and textures. 
This observation also accords with the Fourier analysis in Section \ref{sec:ablation}, which demonstrates that ViT has the characteristics of capturing low-frequency information, and our ViT-Adapter can supplement the missing high-frequency signals.



\textbf{Comparison with SETR.}
Like ViTDet~\citep{li2022exploring}, SETR~\citep{zheng2021rethinking} also changes the shape of features of ViT according to the task prior (see Figure~\ref{fig:comparison}(a)), thus allowing ViT to achieve better segmentation performance. Although this paradigm shares some similarities with our approach, \eg~combining ViT and convolutions, they have three main differences:
(1) In addition to the task prior, our method also takes the information of the input image (the input prior) into consideration when adapting ViT to dense prediction tasks;
(2) The input prior will constantly interact with ViT's features, making the output features more suitable for dense prediction tasks;
(3) Our method is an adapter that is general in both detection and segmentation tasks, and moreover achieves better results than segmentation-specific head SETR~\citep{zheng2021rethinking}.

\textbf{Comparison with other Adapters.} 
We would like to clarify the differences between ViT-Adapter and other adapters~\citep{jia2022visual,bahng2022exploring,chen2022adaptformer,zhang2022neural,jie2022convolutional} for ViTs, from two aspects as follows:

(1) \emph{Different tasks}. Our method is designed for dense prediction tasks, while VPT \citep{jia2022visual}, Visual Prompt \citep{bahng2022exploring}, AdaptFormer \citep{chen2022adaptformer}, NOAH \citep{zhang2022neural}, and Convpass \citep{jie2022convolutional} are mainly proposed for classification tasks. 
By training the parameters only in input spaces, or some modules attached to the backbone, or their combination, these models perform well on classification and even obtain better results than full-tuning models. 

\begin{wraptable}{r}{7cm}\small
\renewcommand\arraystretch{0.9}
\centering
\vspace{-1.3em}
{\setlength{\tabcolsep}{0.6mm}{
    \begin{tabular}[t]{l|c|cc}
    \toprule
           & Trainable  &  mIoU    \\
    Method & Param.     & (ss/ms) \\
    \midrule
     ViT-L (full-tuning) &  318M  & 48.3 / 50.1 \\
     ViT-L (w/ VPT)      & 16M   & 44.0 / 45.6 \\
     \rowcolor{gray!10} 
     ViT-Adapter-L (full-tuning) & 332M & 52.9 / 53.7 \\
     \rowcolor{gray!20} 
     ViT-Adapter-L (frozen ViT) & 30M & 49.0 / 50.6 \\
    \bottomrule
    \end{tabular}}
}
\vspace{-0.5em}
\caption{\textbf{Comparison with VPT.}}
\vspace{-2em}
\label{tab:vpt}
\end{wraptable}
However, when applying these methods~\citep{jia2022visual,bahng2022exploring,chen2022adaptformer,zhang2022neural,jie2022convolutional} to dense prediction tasks, they perform below expectations. For example, we see from Table~\ref{tab:vpt} that the performance of VPT~\citep{jia2022visual} has a large gap with the baseline ViT-L~\citep{zheng2021rethinking} on ADE20K.

(2) \emph{Different targets}. These mentioned adapters~\citep{jia2022visual,bahng2022exploring,chen2022adaptformer,zhang2022neural,jie2022convolutional} aim to explore parameter-efficient transfer learning, while the goal of our ViT-Adapter is to push the performance boundaries of plain ViT downstream applications, make ViT more general for downstream tasks, and efficiently utilize large-scale ViT weights pre-trained in different ways. 
We argue that these two technical lines are orthogonal, as shown in the last column in Table~\ref{tab:vpt}. Combining ViT-Adapter with these adapters to achieve efficient and accurate transfer learning of dense prediction is a research topic worth exploring.

\noindent\textbf{ViTDet's Performance.}
The higher performance of the original ViTDet~\citep{li2022exploring} comes from stronger training settings. Specifically, ViTDet adopts a more expensive training strategy than ours, \ie, loading the MAE~\citep{he2021masked} pre-trained weights, and using the Large Scale Jitter~\citep{ghiasi2021simple} augmentation to train the model for \textbf{100 epochs}. This setting leads to almost 3 times the training cost compared to the commonly used 36 epochs (\ie, $3\times$+MS schedule). And to some extent, it reveals the lack of image-related inductive biases in ViT will lead to slow convergence on dense prediction tasks.

For fair comparisons, we benchmark all plain ViT detectors, including ViTDet~\citep{li2022exploring} and our ViT-Adapter under the commonly used $3\times$+MS training schedule, and use the same ImageNet-1K pre-trained weights (\ie, DeiT) as initialization. It makes sense that our ViT-Adapter achieves better performance than ViTDet under this setting, because our adapter injects image-related prior into the plain ViT, which can speed up convergence and improve performance.

\end{document}